\pdfoutput=1
\documentclass{article}

\usepackage{arxiv}

\usepackage{xcolor}         %

\usepackage{cite}
\usepackage{graphicx}
\usepackage{subcaption}
\usepackage{amsmath}
\usepackage{amsthm}
\usepackage{amssymb}
\newtheorem{definition}{Definition}
\newtheorem{theorem}{Theorem}

\newtheorem{proposition}{Proposition}

\newtheorem{lemma}{Lemma}

\usepackage[linesnumbered,ruled,vlined]{algorithm2e}
\usepackage{subfiles} 
\DeclareMathOperator*{\argmin}{argmin}

\usepackage[utf8]{inputenc} %
\usepackage[T1]{fontenc}    %
\usepackage{hyperref}       %
\usepackage{url}            %
\usepackage{booktabs}       %
\usepackage{amsfonts}       %
\usepackage{nicefrac}       %
\usepackage{microtype}      %
\usepackage{cleveref}       %
\usepackage{lipsum}         %
\usepackage{graphicx}
\usepackage{doi}

\title{Differentially Private Multi-Site Treatment Effect Estimation}

\date{}

\newif\ifuniqueAffiliation
\uniqueAffiliationtrue

\ifuniqueAffiliation %
\author{Tatsuki Koga\\
    Dept. of Computer Science and Engineering\\
    University of California, San Diego\\
    La Jolla, CA\\
    \texttt{tkoga@ucsd.edu}\\
    \And
    Kamalika Chaudhuri\\
    Dept. of Computer Science and Engineering\\
    University of California, San Diego\\
    La Jolla, CA\\
    \texttt{kamalika@cs.ucsd.edu}\\
    \And
    David Page\\
    Dept. of Biostatistics\\
    Duke University\\
    Durham, NC\\
    \texttt{david.page@duke.edu}\\
}
\else
\usepackage{authblk}

\newbox{\orcid}\sbox{\orcid}{\includegraphics[scale=0.06]{orcid.pdf}} 
\author[1]{%
	\href{https://orcid.org/0000-0000-0000-0000}{\usebox{\orcid}\hspace{1mm}David S.~Hippocampus\thanks{\texttt{hippo@cs.cranberry-lemon.edu}}}%
}
\author[1,2]{%
	\href{https://orcid.org/0000-0000-0000-0000}{\usebox{\orcid}\hspace{1mm}Elias D.~Striatum\thanks{\texttt{stariate@ee.mount-sheikh.edu}}}%
}
\affil[1]{Department of Computer Science, Cranberry-Lemon University, Pittsburgh, PA 15213}
\affil[2]{Department of Electrical Engineering, Mount-Sheikh University, Santa Narimana, Levand}
\fi

\renewcommand{\shorttitle}{Differentially Private Multi-Site Treatment Effect Estimation}

\begin{document}
\maketitle

\begin{abstract}
Patient privacy is a major barrier to healthcare AI. For confidentiality reasons, most patient data remains in silo in separate hospitals, preventing the design of data-driven healthcare AI systems that need large volumes of patient data to make effective decisions. A solution to this is collective learning across multiple sites through federated learning with differential privacy. However, literature in this space typically focuses on differentially private statistical estimation and machine learning, which is different from the causal inference-related problems that arise in healthcare. In this work, we take a fresh look at federated learning with a focus on causal inference; specifically, we look at estimating the average treatment effect (ATE), an important task in causal inference for healthcare applications, and provide a federated analytics approach to enable ATE estimation across multiple sites along with differential privacy (DP) guarantees at each site. The main challenge comes from site heterogeneity---different sites have different sample sizes and privacy budgets. We address this through a class of per-site estimation algorithms that reports the ATE estimate \emph{and} its variance as a quality measure, and an aggregation algorithm on the server side that minimizes the overall variance of the final ATE estimate. Our experiments on real and synthetic data show that our method reliably aggregates private statistics across sites and provides better privacy-utility tradeoff under site heterogeneity than baselines.
\end{abstract}

\section{Introduction}
Patient privacy is a major barrier to healthcare AI. Patient confidentiality reasons prevent hospitals and healthcare providers from freely sharing data; consequently, valuable data often remains in silo in separate sites, preventing the development of healthcare AI systems that can learn from large volumes of patient data to make effective decisions. A potential solution to this challenge is collaborative privacy-preserving learning across multiple sites through federated learning with differential privacy. While this has been well-explored for statistical estimation and machine learning problems, these are quite different from the causal inference-related problems that arise in healthcare applications.

This work takes a fresh look at federated learning with differential privacy, and applies it to causal inference---specifically to average treatment effect (ATE) estimation. Here, we are given data $(X_i, Y_i, W_i)$ for patient $i$, where $W_i$ corresponds to a treatment (for example, surgery or not), $Y_i$ to an outcome (for example, recovery or not), and $X_i$ to some covariates or features that describe the patient. The goal is to find the average treatment effect or ATE, which measures if treatment results in a different outcome than non-treatment. While this is easy if the treatments are randomly assigned (that is, under randomized control trials or RCTs), the problem is more challenging with observational data where the assignment of the treatment might depend on the covariates. For example, sicker patients may be denied surgery, which may make surgery look like a more appealing option.

Specifically, we consider the problem of ATE estimation from multiple sites, with differential privacy~\cite{dwork_our_2006-1,dwork_calibrating_2006}, which has emerged as the gold standard in privacy-preserving data analysis. We ensure that each site calculates a DP statistic on its data to ensure the confidentiality of its patients; these statistics are then aggregated by a central server to form an effective ATE estimate.

There are three main challenges in multi-site DP causal inference.
First, for observational studies, where the treatment assignment is not controlled, designing even a single-site DP ATE estimator is not straightforward, and little is known about the problem. In particular, the matching estimator, one of the most standard estimators~\cite{stuart_matching_2010-1}, is hard to sanitize since it can significantly depend on a single individual's data in the worst case.
Second, the estimate quality can vary across sites due to varying sample sizes and privacy budgets. Therefore, each site needs to report not only the ATE estimate but also a quality measure---which sets the problem apart from standard differentially private federated learning estimation solutions. Third, given the ATE estimates and their quality measures, the central server needs to aggregate them appropriately into a final accurate estimate.  

We address the first challenge by proposing a smooth-sensitivity-based DP matching algorithm, SmoothDPMatching. Our algorithm adds significantly less noise for typical real-world datasets than the naive global sensitivity baseline, achieving a better privacy-utility tradeoff.
To deal with the second challenge, we let each site send its ATE estimate variance as a quality measure. Since a site estimates the variance with sensitive data, it publishes the private variance estimate to the server to guarantee privacy.
To address the third challenge, we propose a minimum-variance aggregation algorithm, MVAgg. MVAgg chooses a subset of sites to aggregate so that the variance of the final ATE estimate is minimum.
Combining these three key components gives us a complete method for multi-site DP causal inference.

We evaluate our method on real and synthetic randomized trial and observational study datasets and find that our algorithms lead to significant gains in privacy-accuracy tradeoffs. Specifically, MVAgg automatically adopts estimates from high-quality sites and outperforms baselines, while reliably aggregating the per-site estimates with varying privacy budgets. We also see that SmoothDPMatching considerably reduces the noise variance, and achieves an improved privacy-accuracy tradeoff on both real and synthetic datasets.

\subsection{Related Work}
The most closely related model to our work is the distributed differential privacy model, e.g.,~\cite{dwork_our_2006-1, shi_privacy-preserving_2011-1, bittau_prochlo_2017-1, cheu_distributed_2019-1}, where clients report differentially private output to the untrusted central server.
There has been a body of work on the combination of distributed differential privacy and secure aggregation~\cite{acs_i_2011,goryczka_comprehensive_2017,rastogi_differentially_2010-1, shi_privacy-preserving_2011-1}. Secure aggregation ensures that the server obtains the aggregate result but never sees the individual values. To prevent privacy leakages due to the aggregate result, the clients output locally differentially private (LDP) statistics. 
The fact the server only sees the aggregate result generally amplifies the final central DP guarantee.
Shuffling model is another model of distributed differential privacy~\cite{bittau_prochlo_2017-1, balle_privacy_2019-1, chen_distributed_2021, erlingsson_amplification_2019, cheu_distributed_2019-1, ghazi_scalable_2019-1, ghazi_private_2020, ghazi_private_2020, ghazi_pure_2020}. The model assumes an entity called shuffler, which receives LDP outputs from clients, uniformly permutes them, and sends the shuffled one to the central server. Shuffling further amplifies the privacy guarantee by making it harder for the server to identify individual information.
While these two distributed differential privacy models mainly address privacy amplification on the final central DP guarantee,
our work focuses on how to aggregate client statistics with different qualities to obtain a more accurate final output by the server.

Another line of related work is causal inference under privacy guarantees. The main focus of such papers is to carry out causal inference with privacy in a central DP model, i.e., at a single site, whereas we investigate how causal inference can be done with multiple sites while preserving privacy at each site.
\cite{lee_privacy-preserving_2019} provide a private version of the inverse probability weighting (IPW) method for observational study data.
\cite{kusner_private_2016} study a private procedure to determine whether $X$ causes $Y$ or $Y$ causes $X$ under an additive noise model by privatizing the statistical dependence scores such as Spearman's $\rho$ and Kendall’s $\tau$.
\cite{xu_differential_2017} address private causal graph discovery for categorical and numerical data.
More recently, \cite{niu_differentially_2022} propose a DP meta-algorithm which estimates conditional ATE (CATE).
\cite{komarova_identification_2022} investigate how introducing DP impacts the identification of statistical models.
Note that, to the best of our knowledge, no work has addressed the matching estimator under DP even for a single site setting, which is one of our contributions.

Apart from the privacy literature, there has been a line of work discussing how ATE estimation can be done in multisite random trials under site variation in treatment effect~\cite{fleiss_analysis_1986, raudenbush_statistical_2000, kraemer_pitfalls_2000, weinberger_multisite_2001, weiss_how_2017, robertson_center-specific_2021, dong_design_2021}.
As for learning from data with variate quality, \cite{crammer_learning_2005} provide a theory for choosing an appropriate set of data sources with variable qualities. \cite{song_learning_2015} study how heterogeneous noise impacts the performance of stochastic gradient descent (SGD).

\section{Preliminaries \& Problem Setting}
\label{sec:prelim}
\subsection{Differential Privacy \& Federated Learning/Analytics}

Differential privacy is a strong cryptographically-motivated definition of individual-level privacy. It guarantees that the participation of a single individual in a dataset does not change the probability of any outcome by much. 
In particular, suppose we have two datasets $D$ and $D^\prime$, each consisting of private data from $n$ individuals. We say that $D$ and $D^{\prime}$ are neighboring if they differ in a single individual's private data, i.e., $d(D,D^\prime) = |\{i: D_i\neq D^\prime_i\}| = 1$. The output distribution of a differentially private (randomized) algorithm is guaranteed to be close on neighboring datasets.

\begin{definition}[$(\epsilon, \delta)$-Differential Privacy \cite{dwork_our_2006-1}]
A randomized algorithm $M$ satisfies $(\epsilon, \delta)$-differential privacy if for any two neighboring datasets $D,D^{\prime}$ and for any $S\subseteq \mathrm{range}(M)$,
\begin{align*}
    \Pr [M(D) \in S] \leq \exp (\epsilon) \Pr [M(D^\prime) \in S] + \delta.
\end{align*}
\end{definition}

The most common differentially private mechanism is the Global Sensitivity method, where we compute a function $f$ on a dataset $D$, and add noise that is calibrated to the global sensitivity of the function. Specifically, the global sensitivity of a function $f$ is the maximum difference between the outputs of $f$ on {\em{any two}} neighboring datasets.
The standard instances of the global sensitivity method are the Laplace mechanism~\cite{dwork_calibrating_2006}, which guarantees $(\epsilon,0)$-DP, and the Gaussian mechanism~\cite{dwork_algorithmic_2014}, which guarantees $(\epsilon,\delta)$-DP.

\paragraph{Global Sensitivity \& Laplace~\cite{dwork_calibrating_2006} and Gaussian~\cite{dwork_algorithmic_2014} mechanism.}
The global sensitivity of a scalar function $f:\mathcal{X}^n \to \mathbb{R}$ is 
\begin{align*}
\Delta_{f} = \max_{D,D^\prime} |f(D) -f(D^\prime)|,
\end{align*}
where $D$ and $D^\prime$ are neighboring datasets.\\
Let $\epsilon > 0$ be arbitrary and $f:\mathcal{X}^n \to \mathbb{R}$ be a function. 
Then, the algorithm $M$: 
$M(D) = f(D) + \xi$ 
satisfies $(\epsilon, 0)$-DP, where $\xi\sim\mathrm{Lap}(\nicefrac{\Delta_f}{\epsilon})$. \\
Furthermore, let $\epsilon,\delta \in (0,1)$ be arbitrary and $f:\mathcal{X}^n \to \mathbb{R}$ be a function. 
Then, for $c^2 > 2\ln(1.25/\delta)$, the algorithm $M$:
$M(D) = f(D) + \xi$
satisfies $(\epsilon, \delta)$-DP,
where $\xi\sim \mathcal{N}(0, \sigma^2)$ and $\sigma \geq \frac{c\Delta_{2,f}}{\epsilon}$.

For certain functions, such as the median~\cite{nissim_smooth_2007}, the global sensitivity may be too high, which may lead to a poor privacy-accuracy tradeoff. In these cases, \cite{nissim_smooth_2007} propose calibrating the noise instead to the smoothed sensitivity, which is a smoothed version of the local sensitivity. 
Adding the Laplace noise calibrated to the smooth sensitivity still guarantees DP with a slight overhead in the $\delta$ term.

\paragraph{Local and Smooth Sensitivity \& Laplace mechanism~\cite{nissim_smooth_2007}.}
The local sensitivity of a function $f:\mathcal{X}^n\to \mathbb{R}$ at $D$ is 
\begin{align*}
\mathrm{LS}_f(D) = \max_{D^\prime: d(D,D^\prime)=1} |f(D) -f(D^\prime)|.
\end{align*}
For $\beta > 0$, the $\beta$-smooth sensitivity of $f$ is 
\begin{align*}
  S^*_{f,\beta}(D) = \max_{D^\prime \in \mathcal{X}^n} \mathrm{LS}_f(D^\prime)\cdot \exp(-\beta d(D,D^\prime)).
\end{align*}
If $\beta \leq \nicefrac{\epsilon}{2\ln(\frac{2}{\delta})}$ and $\delta \in (0,1)$, the algorithm $M: \mathcal{X}^n\to \mathbb{R}$:
\begin{align*}
M(D) = f(D) + \frac{2S^*_{f,\beta}(D)}{\epsilon}\cdot \eta,
\end{align*}
where $\eta \sim \mathrm{Lap}(0,1)$, satisfies $(\epsilon,\delta)$-DP.

Federated Learning/Analytics (FL/FA)~\cite{kairouz_advances_2021-2} is an emerging paradigm for collaborative learning across multiple devices or sites, which allows a server to learn a model or some target statistics over sensitive client data, without directly acquiring raw data from the clients. 
However, it is well-known that FL/FA by itself does not directly offer privacy, since the client updates themselves can be reverse-engineered to extract user data~\cite{hitaj_deep_2017,zhu_deep_2019,nasr_comprehensive_2019,wang_beyond_2019,ma_safeguarding_2020-1}. Hence, we will be considering FL/FA with differential privacy. Additionally, we consider FL/FA over a small number of clients, each of which holds data from a certain number of individuals.

\subsection{Average Treatment Effect} \label{sec:prelim-ate}
Suppose we have a group of people who are given a treatment, and our goal is to determine whether the treatment is effective. 
This is done through estimating the Average Treatment Effect (ATE). 
In particular, for an individual $i$, we assume two potential outcomes $Y_i(1)$ and $Y_i(0)$, where $Y_i(1)$ is under treatment and $Y_i(0)$ is under control. 
The average treatment effect (ATE) is then measured by: 
\begin{align*}
\tau = \mathbb{E}[Y_i(1) - Y_i(0)].
\end{align*}

In practice, estimating ATE is not straightforward since we get to observe only one of $Y_i(1)$ and $Y_i(0)$ and cannot directly compute \emph{individual} treatment effect, $Y_i(1) - Y_i(0)$.
Instead, we observe the treatment indicator $W_i$ ($1$ when treated, $0$ under control), the corresponding outcome $Y^\mathrm{obs}_i$, and some other covariates $X_i$.
Then, we aim to estimate ATE given a dataset composed of $N$ individuals' data, $D = \{W_i, Y_i^\mathrm{obs}, X_i\}_{i=1}^{N}$.

We follow the standard causal inference literature \cite{imbens_causal_2015} to make following three assumptions on these variables.
\begin{enumerate}
    \item Stable Unit Treatment Value Assumption (SUTVA): the potential outcomes $(Y_i(1), Y_i(0))$ do not depend on treatments assigned to other individuals
    \item Unconfoundedness: 
    \begin{align*}
    \Pr[W_i=1|X_i, Y_i(1),Y_i(0)] = \Pr[W_i=1|X_i]
    \end{align*}
    \item Positivity: 
    \begin{align*}
    \forall x.\quad 0<\Pr[W_i=1|X_i=x] < 1
    \end{align*}
\end{enumerate}

\subsubsection{Randomized trial and Difference-in-means Estimator} 
In a randomized trial, where treatment assignment is completely random, we estimate ATE via the difference-in-means estimator. Specifically, let $p= \Pr[W_i] (= \Pr[W_i|X_i])$ be some known and controlled assignment probability.
Then, the ATE estimate is: 
\begin{align*}
    \hat{\tau} = \frac{1}{N_t}\sum_{i:W_i=1} Y_i^\mathrm{obs} - \frac{1}{N_c}\sum_{i: W_i=0} Y_i^\mathrm{obs},    
\end{align*}
where $N_t = \sum_i W_i = Np$ and $N_c = \sum_i 1-W_i = N(1-p)$.

\subsubsection{Observational Studies and Matching Estimator}
In observational studies, where the treatment assignment is not controlled, standard practice is to use a matching estimator.
This estimator first imputes the unobserved outcome of an individual by the observed outcome of a \emph{similar} individual who has the opposite treatment status, and then outputs the average of the individual treatment effects.

Among its variants, we use exact single matching under the assumption that there always exists a \emph{similar} individual, i.e.,
for an individual $i$, there exists at least one individual $j$ s.t. $W_i \neq W_j$ and $X_i = X_j$.
Let $m:[N]\to [N]$ be a matching function s.t. $m(i) = j\implies W_i \neq W_j \land X_i = X_j$.
Then, ATE is estimated by:
\begin{align*}
    \hat{\tau} = \frac{1}{N}\sum_{i=1}^N (\hat{Y}_i(1) - \hat{Y}_i(0)),
\end{align*}
where $\hat{Y}_i(1) = W_iY_i^\mathrm{obs} + (1-W_i)Y_{m(i)}^\mathrm{obs}$, and $\hat{Y}_i(0) = W_iY_{m(i)}^\mathrm{obs} + (1-W_i)Y_i^\mathrm{obs}$.
One of $\hat{Y}_i(1)$ and $\hat{Y}_i(0)$ is exactly $Y_i^\mathrm{obs}$ and the other is imputed outcome by the matched individual.

\subsection{Problem Setting}\label{sec:problem-setting}
\begin{figure}[t]
    \centering
\minipage{0.48\textwidth}
    \includegraphics[width=0.95\linewidth]{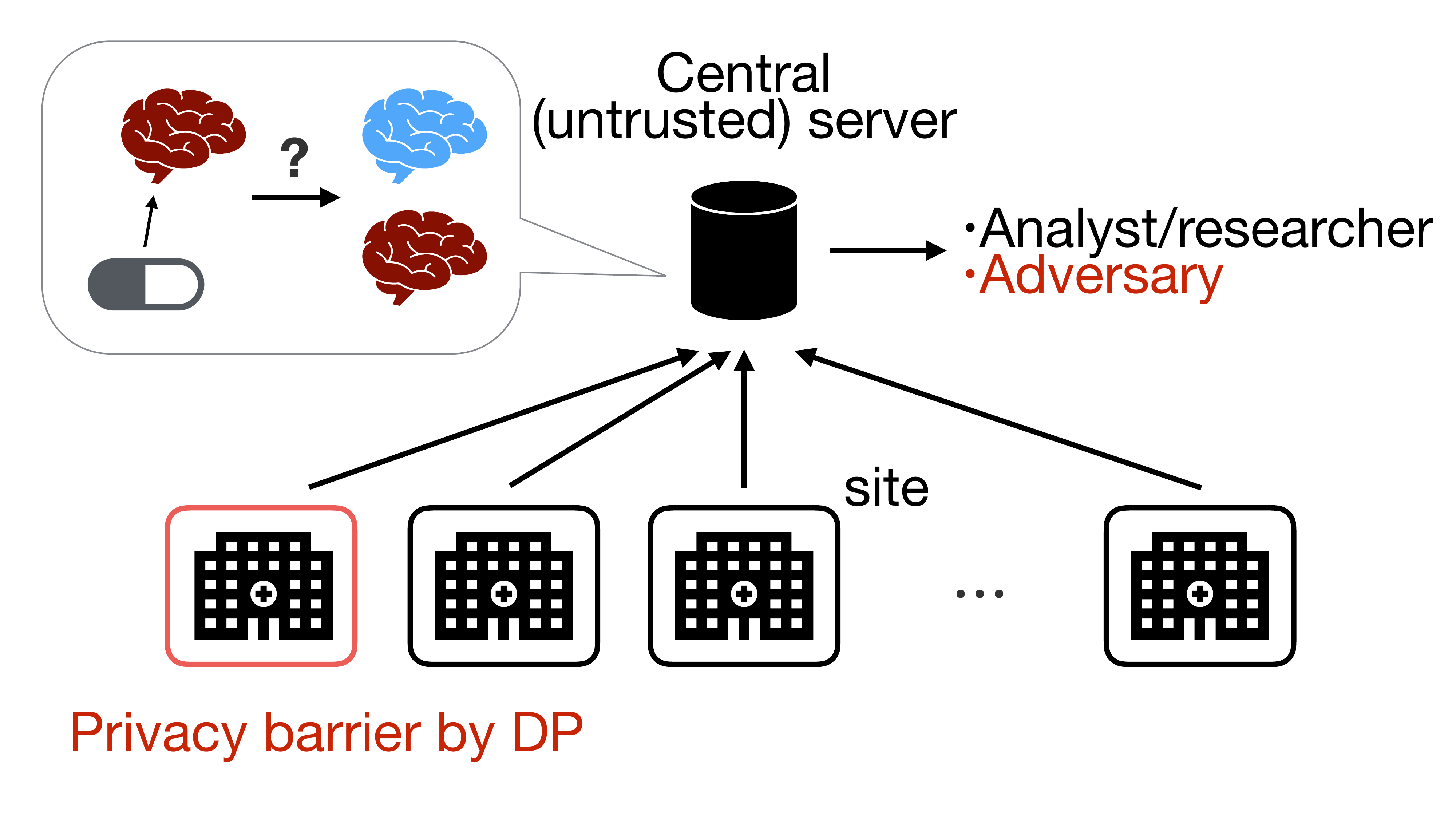}
    \caption{Our framework on estimating ATE with data from distributed sites}
    \label{fig:framework}
\endminipage\hfill
\minipage{0.48\textwidth}
    \centering
    \includegraphics[width=0.95\linewidth]{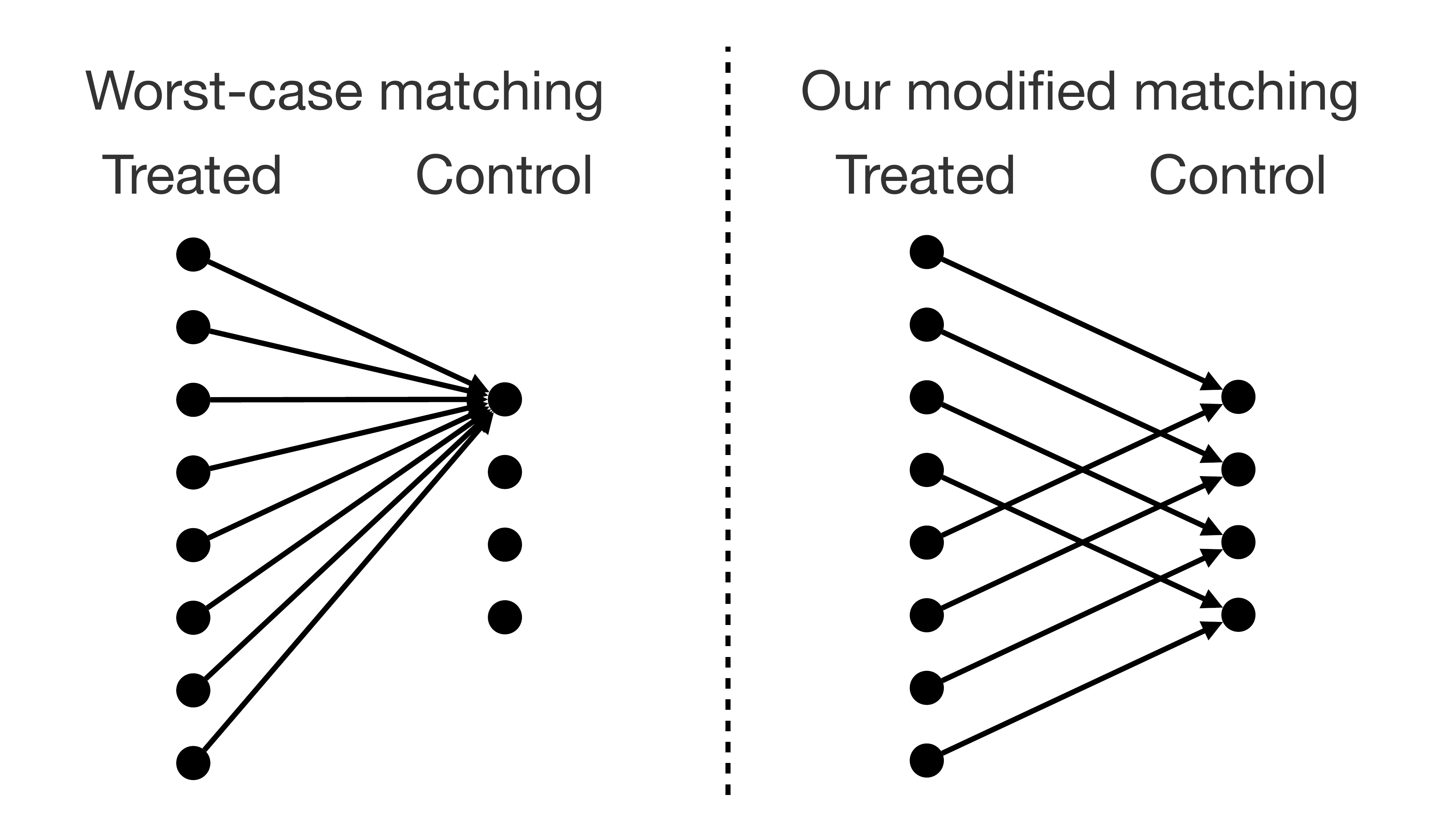}
    \caption{Worst-case matching (left) and our modified matching (right) within the same covariate stratum. We omit arrows from control individuals to treated ones for readability.}
    \label{fig:exact-matching}
\endminipage
\end{figure}

Our goal is to estimate the ATE of a specific binary treatment, where data about the effect of this treatment is distributed across a small number of sites. 
We would like to ensure that raw data stays on the site, and only differentially private estimates leave a particular site. 
Specifically, the ATE computation is done by an untrusted server, which receives private statistics from $J$ sites. 
We demonstrate the figure for this framework in Figure~\ref{fig:framework}.

\paragraph{Some Basic Notation.} 
We assume that site $j$ requires $(\epsilon_j, \delta_j)$-DP, and use the notation $\hat{\tau}_{j-\mathrm{DP}}$ to denote the ATE at site $j$. 
The final ATE estimated at the server is denoted by $\hat{\tau}_\mathrm{DP}$.
Furthermore, site $j$ has a dataset $D_j$ of size $N_j$.
Note that the sample sizes, $N_j$'s, are the public information since we are interested in the site-level privacy guarantee.
$i$-th element in $D_j$ is a tuple $(W_{ij}, Y_{ij}^\mathrm{obs}, X_{ij})$, 
where $W_{ij} \in \{0,1\}$ is the status of binary treatment, $Y_{ij}^\mathrm{obs}$ is the observed outcome, which is assumed to be bounded, i.e., $0 \leq Y_{ij}^\mathrm{obs} \leq B$ for some $B$, and $X_{ij}$ is the covariates, if any, of $i$-th individual at site $j$.
$X_{ij}$ is used for observational studies and is typically a vector of multiple covariates.
In this work, we assume that it is an element of some finite set $\mathcal{X}$.
In practice, covariates can contain continuous values, e.g., height and weight, but we can often discretize them without losing much precision.

\paragraph{Assumptions.} 
In addition to the three standard causal inference assumptions~\cite{imbens_causal_2015}, we make two other mild assumptions. 
The first is that the sites are homogeneous. 
That is, if each individual has potential outcomes, $Y_{ij}(1)$ and $Y_{ij}(0)$, we further assume that a tuple $(W_{ij}, Y_{ij}(1), Y_{ij}(0), X_{ij})$ is drawn i.i.d. from some fixed distribution. 
This assumption is needed so that the estimand $\tau=\mathbb{E}[Y_i(1)-Y_i(0)]$ makes sense; without this, the underlying ATE at each site differs and it is no longer clear what estimand we should use.
Our second assumption is that an individual cannot belong to more than one site. This ensures the privacy loss does not accumulate by outputs from multiple sites.

\section{Method}

Our method consists of two interconnected components on a distributed client-server setting---first, a site-level estimation algorithm and second, a server-side aggregation algorithm. Combining these two components gives us a complete method for private and distributed ATE estimation. 

\subsection{Per-Site Estimation Algorithm}
In a FL/FA setting, a per-site estimation algorithm computes a per-site gradient/target statistic on its local data, adds noise for privacy, and sends it to the server. 
This simple solution, however, does not directly apply to us. 
For the server to aggregate the ATEs appropriately, it needs to know a quality measure for the ATEs from each site because the estimation quality can vary across sites due to varying sample sizes and privacy budgets.
For example, even though all the sites have the same total privacy budget, they can answer multiple queries on the same data, and the privacy budgets allocated for an ATE estimation query could differ by site.
To this end, we calculate a differentially private variance estimate for the ATE, which provides a comprehensive estimate by taking into account non-private estimate variance as well as additive noise for DP.
Another difficulty for us is that, in the common observational data case, standard ATE estimators are more involved than the sum or average over individual values. Therefore, it is not obvious how to construct their DP versions.
Thus, we propose a new smooth-sensitivity-based DP matching algorithm, SmoothDPMatching, through an analysis of the smooth sensitivity of the matching estimator. Our algorithm significantly reduces the noise variance and hence improves the accuracy for typical datasets compared with the baseline global sensitivity method.

\subsubsection{Randomized Trial and Difference-in-means Estimator} \label{sec:per-site-rand-trial}

For randomized trials, we use the difference-in-means estimator for ATE---namely,
$\hat{\tau} = \sum_{i:W_i=1} Y_i^\mathrm{obs}/N_t - \sum_{i: W_i=0} Y_i^\mathrm{obs}/N_c$.
Its differentially private version can be straightforwardly computed using the global sensitivity method.
The global sensitivities of $\sum_{i: W_i=1} Y_i^\mathrm{obs}$ and $\sum_{i: W_i=0} Y_i^\mathrm{obs}$ are both $B$.
Thus, by using the Laplace mechanism, we have:
\begin{align*}
\hat{\tau}_\mathrm{DP} = \frac{1}{N_t}\left(\sum_{i: W_i=1} Y_i^\mathrm{obs} + \xi_t\right) - \frac{1}{N_c}\left(\sum_{i: W_i=0} Y_i^\mathrm{obs} + \xi_c\right)
=\hat{\tau} + \frac{\xi_t}{N_t} - \frac{\xi_c}{N_c},
\end{align*}
where $\xi_t,\xi_c \sim \mathrm{Lap}(\nicefrac{B}{\epsilon_1})$.
By the parallel composition theorem of DP, the mechanism satisfies $\epsilon_1$-DP.

The private variance estimation of $\hat{\tau}_\mathrm{DP}$ is also simple because $\xi_t$ and $\xi_c$ are independent from the data distribution.
That is, we have 
\begin{align*}
\mathbb{V}[\hat{\tau}_{\mathrm{DP}}] = \mathbb{V}[\hat{\tau}] + \mathbb{V}\left[\frac{\xi_t}{N_t}\right] + \mathbb{V}\left[\frac{\xi_c}{N_c}\right]
= \mathbb{V}[\hat{\tau}] + \frac{2B^2 (\frac{1}{N_t^2}+\frac{1}{N_c^2})}{\epsilon_1^2},
\end{align*}
where the last term is computed only with public information.
It remains to estimate the sampling variance term, $\mathbb{V}[\hat{\tau}]$, with sensitive data and sanitize the estimate using the global sensitivity method.
In particular, $\mathbb{V}[\hat{\tau}]$ is estimated with $s^2_t/N_t + s^2_c/N_c$~\cite{imbens_causal_2015}, where $s^2_t$ and $s^2_c$ are sample variance of $\sum_{i: W_i=1} Y_i^\mathrm{obs} / N_t$ and $\sum_{i: W_i=0} Y_i^\mathrm{obs} / N_c$.
It suffices to privately estimate $\sum_{i: W_i=1} (Y_i^\mathrm{obs})^2$ to obtain the private estimate of $s^2_t$ because $s^2_t = \sum_{i: W_i=1} (Y_i^\mathrm{obs})^2 / N_t - (\sum_{i: W_i=1}Y_i^\mathrm{obs} / N_t)^2$ and we already know the private estimate of $\sum_{i: W_i=1} Y_i^\mathrm{obs}$.
The global sensitivity of $\sum_{i: W_i=1} (Y_i^\mathrm{obs})^2$ is $B^2$; thus, we apply the Laplace mechanism with a parameter $\epsilon_2$ and obtain the private estimate of $s^2_t$.
The same argument yields the private estimate of $s^2_c$.
By the parallel composition theorem, this computation satisfies $\epsilon_2$-DP.
Consequently, we obtain the private variance estimate of the differentially private difference-in-means estimator.
We send this estimation along with $\hat{\tau}_\mathrm{DP}$ to the server, which satisfies $(\epsilon_1+\epsilon_2)$-DP by the sequential composition theorem of DP.

\subsubsection{Observational Study and Matching Estimator} \label{sec:per-site-obs}
Things however are more complicated for observational data, since we need to match covariates. To do this privately, we propose a new differentially private approximation to the exact matching estimator in Section~\ref{sec:prelim}. The main challenge here is that changing one value in the input dataset can alter the final output significantly \emph{in the worst case}. We address this through a smooth-sensitivity-based algorithm, SmoothDPMatching, which requires much less noise for typical datasets.

\begin{algorithm*}[tb]
\DontPrintSemicolon
  \KwData{Dataset: $D = \{(W_i, Y_i^\mathrm{obs}, X_i)\}_{i=1}^{N}$}
        Define placeholders $\{(\hat{Y}_i(1),\hat{Y}_i(0))\}_{i=1}^N$\;
        \For{$x \in \mathcal{X}$}{
            $T_x = \{i:W_i=1 \land X_i=x\}$\;
            $C_x = \{i:W_i=0 \land X_i=x\}$\;
            \If(\tcc*[h]{when there's no match}){$|T_x|=0 \lor |C_x|=0$}{
                \For{$i \in T_x \cup C_x$}{$(\hat{Y}_i(1),\hat{Y}_i(0))\leftarrow(Y_i^\mathrm{obs},Y_i^\mathrm{obs})$}
            }
            \Else(\tcc*[h]{when there are matches}){
                \For{$j=0$ \KwTo $|T_x|-1$}{
                    $i \leftarrow T_x[j]$\;
                    $m(i) \leftarrow C_x[j \bmod |C_x|]$ \tcc*[h]{matched individual}\;
                    $(\hat{Y}_i(1),\hat{Y}_i(0)) \leftarrow (Y_i^\mathrm{obs},Y_{m(i)}^\mathrm{obs})$
                }
                \For{$j=0$ \KwTo $|C_x|-1$}{
                    $i \leftarrow C_x[j]$\;
                    $m(i) \leftarrow T_x[j \bmod |T_x|]$ \tcc*[h]{matched individual}\;
                    $(\hat{Y}_i(1),\hat{Y}_i(0)) \leftarrow (Y_{m(i)}^\mathrm{obs},Y_i^\mathrm{obs})$
                }
            }
        }
        Compute non-private ATE: $\hat{\tau}=\frac{1}{N}\sum_{i=1}^N \hat{Y}_i(1) - \hat{Y}_i(0)$\;
        \KwRet $\hat{\tau}$\;
  \caption{Non-private Matching at site}
  \label{alg:non-private-matching}
\end{algorithm*}

\begin{algorithm*}[tb]
\DontPrintSemicolon
  \KwData{Dataset: $D = \{(W_i, Y_i^\mathrm{obs}, X_i)\}_{i=1}^{N}$, Privacy parameters: $\epsilon$, $\delta$}
        $\beta = \frac{\epsilon}{2\ln (\frac{2}{\delta})}$\;
        Compute $S_{\hat{\tau}, \beta}^*(D)$ as in Eq.~\eqref{eq:smooth-sens}\;
        Compute non-private ATE: $\hat{\tau}=\frac{1}{N}\sum_{i=1}^N \hat{Y}_i(1) - \hat{Y}_i(0)$ with Algorithm~\ref{alg:non-private-matching}\;
        $\hat{\tau}_\mathrm{DP} = \hat{\tau} + \frac{2S^*_{\hat{\tau}, \beta}(D)}{\epsilon}\cdot \eta$, where $\eta \sim \mathrm{Lap}(1)$\;
        \KwRet $\hat{\tau}_\mathrm{DP}$\;
  \caption{SmoothDPMatching at site}
  \label{alg:dp-matching}
\end{algorithm*}

We provide the specification to the estimator to make it more amenable to DP. 
First, recall that the exact matching estimator assumes that every individual in the data has an exact match. If an individual $i$ does not have any matched individual $j$, then we extend the matching function $m$ to be $m:[N]\to [N] \cup \{\bot\}$ and set $\hat{Y}_i(1) = \hat{Y}_i(0) = Y_i^\mathrm{obs}$ when $m(i) = \bot$ so that this term contributes $0$ to the ATE. 
Second, for each covariate stratum $X=x$, the exact matching estimator may match a control individual to many treated individuals while other control individuals have no matches, driving up the sensitivity. We ensure that we balance the number of individuals matched to a particular individual for each covariate stratum as shown in Figure~\ref{fig:exact-matching}. 
This can be done with a greedy algorithm shown in Algorithm~\ref{alg:non-private-matching}. 
We provide the further details in Appendix.

Even with the specification, 
unfortunately, it is shown as below that the global sensitivity of the matching algorithm is still a constant. We therefore use a smoothed sensitivity estimator. 
\begin{proposition}
Let $\hat{\tau}$ be the exact single matching estimator in Algorithm~\ref{alg:non-private-matching}, then $\Delta_{1,\hat{\tau}} \geq B$.
\end{proposition}
\begin{proof}
Consider a pair of neighboring datasets $D,D^\prime$ where $|D|=|D^\prime|=N$, $D_N\neq D_N^\prime$, $D_i = (W_i,Y_i^\mathrm{obs},X_i) = (1, B, x)$ for $i=1,\ldots,N-1$, $D_N = (0,0,x)$, and $D_N^\prime = (1,0,x)$.
Here, $D^\prime$ only contains treated individuals; thus, $\hat{\tau}(D^\prime) = 0$.
Then, $|\hat{\tau}(D) - \hat{\tau}(D^\prime)| = |\frac{1}{N} (\sum_{i=1}^{N-1} (Y_i - Y_{m(i)}) + (Y_{m(N)} - Y_N)) - 0| = \frac{1}{N}((N-1)B+B) = B$.
By the definition of the global sensitivity, the statement is shown.
\end{proof}

We first analyze the smooth sensitivity of our exact single matching estimator, $\hat{\tau}$.
For most real datasets, we anticipate the local sensitivity is $\mathcal{O}(1/N)$ whereas the global sensitivity is $o(1)$.
Then, the smooth sensitivity, the smooth upper bound of the local sensitivity, is also $\mathcal{O}(1/N)$.
This is in fact true as stated in the theorem below.

\begin{theorem}
Let $T_x = \{i: W_i=1 \land X_i=x\}$ and $C_x = \{i: W_i=0 \land X_i=x\}$ be the sets of treated and control individuals with the covariate $x$.
Then, the local sensitivity of $\hat{\tau}$ is upper bounded as follows: 
\begin{align*}
    \mathrm{LS}_{\hat{\tau}}(D) \leq 
    \frac{1}{N}\max_{x\in \mathcal{X}: |T_x|>0 \lor |C_x|>0}
    \begin{cases}
    4(1+\max(|T_x|, |C_x|))B & |T_x| = 0 \lor |C_x| = 0\\
    4(1+\max(\lceil \frac{1+|C_x|}{|T_x|}\rceil, \lceil \frac{1+|T_x|}{|C_x|}\rceil))B & o.w.
    \end{cases}.
\end{align*}
Furthermore, let 
$R_x^{(k)}(D) = \begin{cases}
    \max(|T_x|,|C_k|) +k & \min(|T_x|,|C_x|) \leq k\\
    \lceil \frac{\max(|T_x|,|C_k|)+k+1}{\min(|T_x|,|C_x|)-k} \rceil & o.w.\\
    \end{cases}$.
Then, the $\beta$-smooth sensitivity of $\hat{\tau}$ is upper bounded as follows:
\begin{align} \label{eq:smooth-sens}
    S^*_{\hat{\tau},\beta}(D) = \max_{k=0,\ldots, N} e^{-k\beta} \frac{4B}{N}(1+\max_{x\in\mathcal{X}} R^{(k)}_x(D)).
\end{align}
In addition, $S^*_{\hat{\tau},\beta}(D)$ can be computed with $\mathcal{O}(\min(|\mathcal{X}|,N))$ space and $\mathcal{O}(N\cdot \min(|\mathcal{X}|,N))$ time. 
\end{theorem}
We provide the proof in Appendix.

We observe that if the dataset is well-balanced in each covariate value $x$, i.e., $|T_x| \approx |C_x|$, $\mathrm{LS}_{\hat{\tau}}(D) = \mathcal{O}(1/N)$ and also $S^*_{\hat{\tau},\beta}(D)= \mathcal{O}(1/N)$.
In contrast, the global sensitivity is $o(1)$ regardless of a dataset.
We demonstrate this observation by numerical simulations on a synthetic dataset in Section~\ref{sec:exp}.

For completeness, we present a $(\epsilon,\delta)$-DP matching algorithm shown in Algorithm~\ref{alg:dp-matching}, which calibrates the Laplace noise to the analyzed smooth sensitivity.

Next, we turn to estimating the variance of our DP matching estimator privately.
The variance estimate is slightly more involved since the additive noise variance now depends on the smooth sensitivity, which is \emph{data-dependent}.
Thus, we instead obtain the private estimate of the smooth sensitivity, in addition to the sampling variance, to get the overall variance estimate.

More formally, we consider the variance conditioned on $X_i$'s and $W_i$'s as in the literature~\cite{imbens_causal_2015}.
Recall we have $\hat{\tau}_\mathrm{DP} = \hat{\tau} + (2S^*_{\hat{\tau}, \beta}(D)/\epsilon) \cdot \eta$, where $\eta \sim \mathrm{Lap}(1)$ (line 4 in Algorithm~\ref{alg:dp-matching}).
Then, the variance of $\hat{\tau}_\mathrm{DP}$ is:
\begin{align*}
\mathbb{V}[\hat{\tau}_{\mathrm{DP}}] = \mathbb{V}[\hat{\tau}] + \mathbb{V}\left[\frac{2S^*_{\hat{\tau}, \beta}(D)}{\epsilon} \cdot \eta\right]
= \mathbb{V}[\hat{\tau}] + \frac{8(S^*_{\hat{\tau},\beta}(D))^2}{\epsilon^2}.
\end{align*}
It remains to estimate both terms from data privately.
Note that, conditioned on $X_i$'s and $W_i$'s, the smooth sensitivity in eq.~\eqref{eq:smooth-sens} is constant.

As for the sampling variance term, $\mathbb{V}[\hat{\tau}]$, we obtain its differentially private estimate by the smooth sensitivity method.
We defer the detail to Appendix.
As for the second term, $8(S^*_{\hat{\tau},\beta}(D))^2/\epsilon^2$, we need to privately estimate the smooth sensitivity because it is data-dependent.
We thus provide an \emph{unbiased} $(\epsilon,\delta)$-DP estimator of the $\beta$-smooth sensitivity as follows.
Since the smooth sensitivity is the \emph{smoothed} version of the local sensitivity, it is designed not to vary a lot by changing a single individual's data.
Therefore, we first show that the global sensitivity of $\ln S^*_{\hat{\tau},\beta}$ is $\beta$.
Then, we apply the Gaussian mechanism to obtain the differentially private smooth sensitivity, $S^*_{\hat{\tau},\beta-\mathrm{DP}}(D)$.

\begin{lemma}
Let $S^*_{f,\beta}$ be the $\beta$-smooth sensitivity of $f$.
Then, 
\begin{align*}
    S^*_{f,\beta-\mathrm{DP}}(D) = \exp(\ln S^*_{f,\beta}(D) + z - \frac{\sigma^2}{2}),
\end{align*} 
where $\sigma = \sqrt{2\ln (1.25/\delta)}\beta/\epsilon$ and $z\sim \mathcal{N}(0,\sigma^2)$, satisfies $(\epsilon,\delta)$-differential privacy.\\
Furthermore, it holds that $\mathbb{E}[S^*_{f,\beta-\mathrm{DP}}(D)] = S^*_{f,\beta}(D)$, where the randomness is over the draws of $z$.
\end{lemma}
\begin{proof}
    By the definition of smooth sensitivity, it holds that for any neighboring datasets $D,D^\prime$, $S^*_{f,\beta}(D) \leq e^\beta S^*_{f,\beta}(D^\prime)$.
    Therefore, the global sensitivity of $\ln S^*_{f,\beta}$ is $\beta$.
    The privacy guarantee of the Gaussian mechanism and the post-processing theorem of DP guarantee $(\epsilon,\delta)$-DP for $S^*_{f,\beta-\mathrm{DP}}$.

    Additionally, by taking the expectation over the draws of $z$, we have:
    \begin{align*}
        \mathbb{E}[S^*_{f,\beta-\mathrm{DP}}(D)] = S^*_{f,\beta}(D) \cdot \exp(-\frac{\sigma^2}{2})\cdot\mathbb{E}[\exp(z)]
        = S^*_{f,\beta}(D),
    \end{align*}
    where the last equality holds due to $\mathbb{E}[\exp(z)]=\exp(\frac{\sigma^2}{2})$.
\end{proof}

\subsection{Aggregation Algorithm on Server}
\begin{algorithm}[t]
\DontPrintSemicolon
  \KwData{Sample sizes: $N_1,\ldots, N_J$, Noisy estimates and their variances: $\{\hat{\tau}_{j-\mathrm{DP}}, \hat{\sigma}^2_{j-\mathrm{DP}}\}_{j=1}^{J}$}
        $I^* = \argmin_{I \subseteq [J]}\sum_{j\in I}(\frac{N_j}{N_I})^2 \hat{\sigma}^2_{j-\mathrm{DP}}$\;
        $\hat{\tau}_\mathrm{DP} = \sum_{j\in I^*}\frac{N_j}{N_{I^*}} \hat{\tau}_{j-\mathrm{DP}}$\;
        \KwRet $\hat{\tau}_\mathrm{DP}$\;
  \caption{MVAgg at server}
  \label{alg:minimum-variance-agg}
\end{algorithm}

In the simple FL/FA, the server would simply average the gradients/statistics transmitted by the clients. 
Unfortunately for us, this solution is not enough---different sites will have different estimation quality, due to varying dataset size and/or varying privacy budgets.
We therefore propose a new aggregation procedure that takes this heterogeneity into account. 

Since we are interested in the average treatment effect on \textit{an individual}, we consider the weighted average of ATEs from sites with weights proportional to the sample sizes at sites $N_j$'s.
Given a set of sites $I$ and a set of DP ATE estimates $\{\hat{\tau}_{j-\mathrm{DP}}\}_{j\in I}$, let $N_I = \sum_{j\in I} N_j$, and the server publishes
$\hat{\tau}_\mathrm{DP} = \sum_{j\in I}(\nicefrac{N_j}{N_I})\cdot \hat{\tau}_{j-\mathrm{DP}}$.

The central problem at the server is then how to choose the set of sites $I$.
When some sites in $I$ have very noisy ATE estimates, the final estimate can be noisy, or has high variance, as well.
In such a case, we might want to remove these sites from the set so that the final estimate is less noisy.
Therefore, we propose a new aggregation algorithm that embodies this idea by choosing the set of sites that minimizes the variance of the aggregate ATE: minimum-variance aggregation algorithm (MVAgg).

More concretely, the minimum-variance aggregation algorithm shown in Algorithm~\ref{alg:minimum-variance-agg} takes noisy ATEs, noisy variance of ATE, and sample sizes as the inputs.
Then, it minimizes the estimated variance over a set of sites.
Here, since $\hat{\tau}_\mathrm{DP}$ is the weighted average of $\hat{\tau}_{j-\mathrm{DP}}$'s, its variance given the set $I$ is $\mathbb{V}[\hat{\tau}_\mathrm{DP}] = \sum_{j\in I}(N_j/N_I)^2 \mathbb{V}[\hat{\tau}_{j-\mathrm{DP}}]$.
It finally computes the weighted average of the noisy ATEs over the chosen set of sites.
Note that by the post-processing theorem of DP, the privacy guarantee at each site never changes as a result of the aggregation.

Our algorithm is general in the sense that it only requires the noisy estimate and its noisy variance from each site in addition to the publicly known sample sizes,
and it does not limit the specific estimator used at each site.
On the other hand, our algorithm currently adopts a brute-force search to determine the minimum variance set (line 1 in Algorithm~\ref{alg:minimum-variance-agg}).
Finding a greedy approximation algorithm for the minimization is a possible direction for our future work.

\section{Experiment}\label{sec:exp}
We now empirically investigate how putting together our estimation algorithms at the client sites with the aggregation process on the server side works. Specifically, we ask the following questions:
\begin{enumerate}
    \item How does the smooth-sensitivity-based DP matching algorithm (Algorithm~\ref{alg:dp-matching}) improve the privacy-utility tradeoff on observational study data at each site?
    \item How does our aggregation algorithm (Algorithm~\ref{alg:minimum-variance-agg}) impact the final ATE estimation on the server on randomized trial and observational study data?
    \item How do site-level privacy parameters affect the overall performance of the algorithms?
\end{enumerate}

We answer the first question with real and synthetic observational study data. We then answer the rest of the questions with real randomized trial data as well as those observational data.

\subsection{Methodology}
\paragraph{Datasets.} 
For randomized trial data, we use two real datasets. The International Stroke Trial (\textbf{IST})~\cite{noauthor_international_1997} is a dataset with $N=18995$ individuals, where $N_t=9705$ are randomly treated by the aspirin allocation and $N_c=9703$ are controlled. The outcome measures whether the recurrent ischemic stroke occurs within 14 days after treatment. 
Tennessee's Student Teacher Achievement Ratio (\textbf{STAR}) dataset~\cite{word_studentteacher_1990} contains the trial results from $N=10331$ students, who are randomly assigned into either a small class ($N_t = 2643$) or regular-size class ($N_c = 7688$).
The data is collected from 80 schools. We use the four kinds of school urbanity (rural, suburban, urban, inner city) to determine which site the student belongs to, i.e., $J=4$. The sample sizes result in $N_1:N_2:N_3:N_4 \approx 5:3:3:1$.
\\
For observational study data, we use a synthetic, a semi-real, and a real dataset.
We generate the synthetic dataset (\textbf{Synth}) by first sampling $X_i$'s uniform randomly from a discrete set $\mathcal{X}=\{0,1/(|\mathcal{X}|-1),\ldots,1\}$. 
Then, we sample $W_i$ from the Bernoulli distribution with a parameter $\mathrm{sigmoid}(a\cdot (2X_i-1))$ with some $a\in\mathbb{R}$, where a parameter $a$ is drawn from $\mathrm{Uniform}([-1,1])$ if not mentioned or set as constant. This ensures that the treatment variable $W_i$ has some dependence on $X_i$. Note that as $a$ gets larger, the dataset gets more imbalanced, i.e., $|T_x| \gg |C_x|$ or $|T_x| \ll |C_x|$. Finally, given the underlying true ATE $\tau=0.5$, we set $Y_i=b\cdot X_i + \tau\cdot W_i + e_i$, where $b\sim \mathrm{Uniform}([0,0.4])$ is a parameter and $e_i \sim \mathrm{Uniform}([0,0.1])$ is observation noise.
This generation process ensures that $Y_i$ depends on both $X_i$ and $W_i$ and that $0\leq Y_i \leq 1$.
The semi-real dataset we use is the Infant Health and Development Program (\textbf{IHDP}) dataset~\cite{hill_bayesian_2011}, where only the outcome value is simulated.
It has $N=747$ individuals comprised of $N_t=139$ treated and $N_c=608$ controlled individuals.
The treatment is specialist home visits to the children and the outcome is future cognitive test scores.
We choose 3 discrete covariates out of 25 covariates in the original dataset to ensure the exact matching.
The real dataset we use is \textbf{Lalonde}~\cite{lalonde_evaluating_1986}, which is composed of $N=722$ individuals where $N_t=297$ are treated and $N_c=425$ are controlled.
The treatment is job training and the outcome is earning in 1978.
We choose age as the only covariate to ensure the exact matching.
For all datasets, we preprocess them so that $0\leq Y_i\leq 1$.

\paragraph{Algorithms.}
The per-site estimation algorithms used in the experiments are as follows.
We use the DP version of the difference-in-means estimator presented in Section~\ref{sec:per-site-rand-trial} for randomized trial data.
For observational study data, we compare two DP matching algorithms whose additive noises are calibrated to the global sensitivity (\textbf{GlobalDPMatching}) and smooth sensitivity (Algorithm~\ref{alg:dp-matching}; \textbf{SmoothDPMatching}) respectively. 
\\
We compare three aggregation algorithms on the server: aggregation of results from all sites (\textbf{AggAll}), publishing the result of the largest site (\textbf{AggLargest}), and \textbf{MVAgg} (Algorithm~\ref{alg:minimum-variance-agg}).

\paragraph{Experiment Setup.}
We consider two-site ($J=2$) and three-site ($J=3$) settings on the datasets except for STAR dataset where the sites are pre-assigned ($J=4$). 
For the two-site setting, we randomly assign individuals to each site while keeping the sample sizes equal, $N_1=N_2$~\footnote{We assume AggLargest always chooses the first site as the largest site.}.
For the three-site setting, we consider different sample size proportions as follows:
$N_1:N_2:N_3 = 1:1:1$, $3:2:1$, $9:9:2$, and $18:1:1$. 
\\
We fix the privacy parameter for the first site, $\epsilon_1$, and sweep the others, $\epsilon_2,\ldots, \epsilon_J$. In particular, for each $\alpha \in \{1/8, 1/4, 1/2, 1, 2, 4, 8\}$, let $\epsilon_j = \alpha^{(j-1)/(J-1)} \epsilon_1$. 
As $\alpha$ gets larger, the second to $J$-th sites are expected to send more accurate statistics.
We use $\epsilon_1=1$ for IST, STAR, and Synth, with $N=10000$ and $|\mathcal{X}|=100$, and $\epsilon_1=5$ for IHDP and Lalonde due to their small sample sizes.
We further fix $\delta_1=\cdots = \delta_J=10^{-5}$.
For the per-site estimation, we evenly split the privacy budget into multiple DP algorithms, e.g., we assign $(\epsilon/3, \delta/3)$ separately for obtaining the ATE estimate, the sampling variance estimate, and the private smooth sensitivity.
\\
The evaluation metric is the mean absolute error (MAE) between the non-private and private ATE estimates. For Synth data, we measure the MAE between the true underlying ATE and the private estimate. We repeat the algorithms 100 times and report the mean and standard deviation of MAE.

\subsection{Results}
\subsubsection{Randomized Trial and Difference-in-means Estimator}
\begin{figure}[t]
  \centering
  \minipage{0.48\textwidth}
  \includegraphics[width=0.9\linewidth]{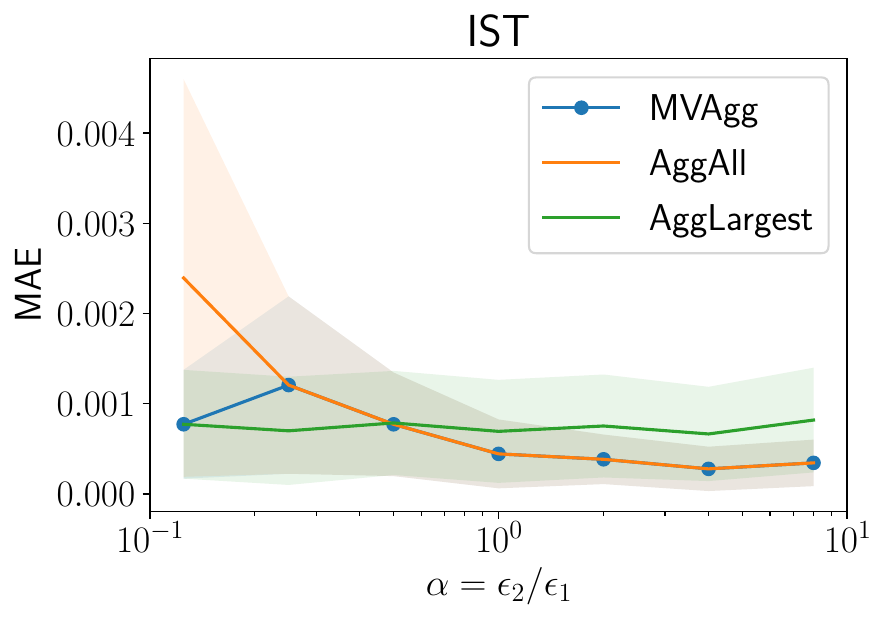}
  \caption{Mean MAEs and standard deviations of MVAgg, AggAll and AggLargest on IST dataset under two-site setting}\label{fig:ist-two-varying-eps}
  \endminipage\hfill
  \minipage{0.48\textwidth}
  \includegraphics[width=0.9\linewidth]{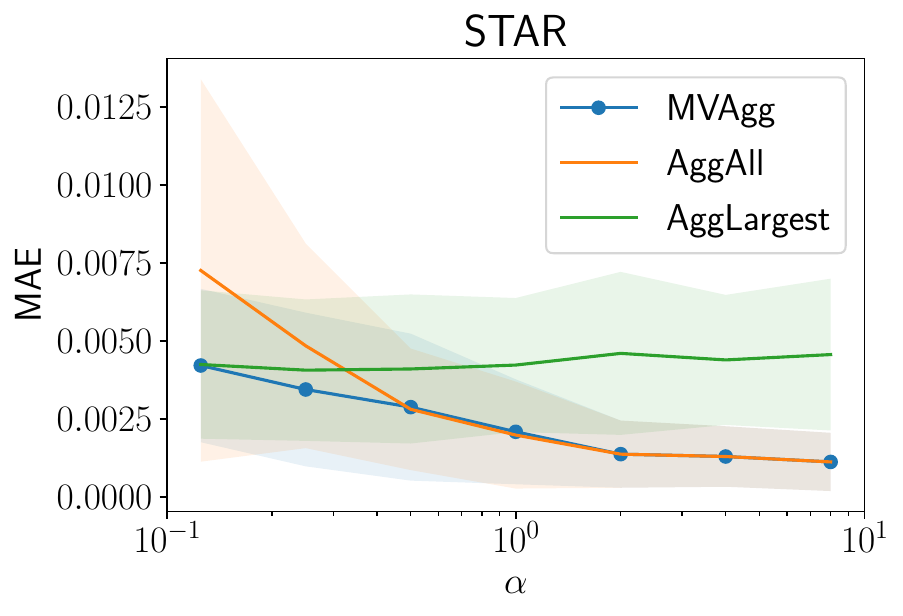}
  \caption{Mean MAEs and standard deviations of MVAgg, AggAll and AggLargest on STAR dataset ($J=4$)}\label{fig:star-four-varying-eps}
  \endminipage
\end{figure}
\begin{figure}[t]
\begin{subfigure}{.24\textwidth}
  \centering
  \includegraphics[width=\linewidth]{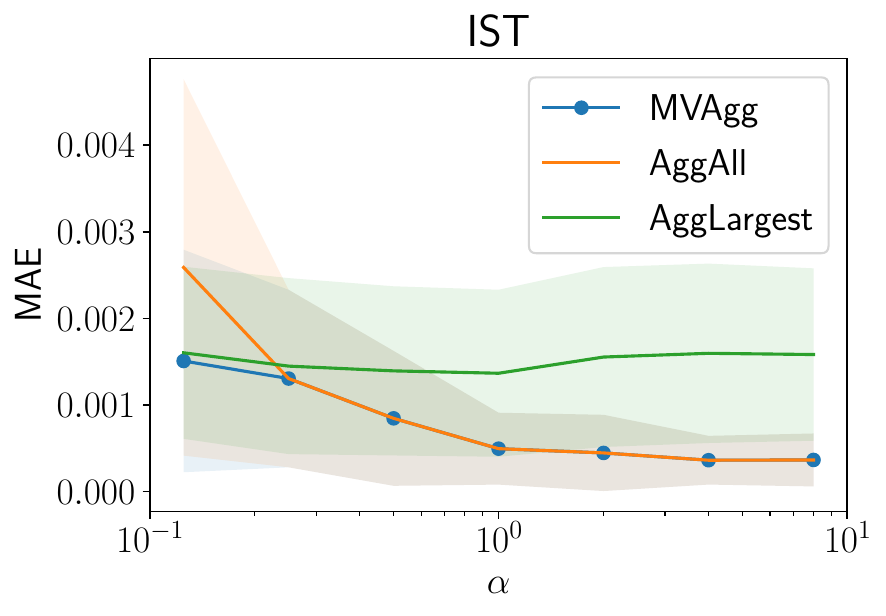}
\end{subfigure}%
\begin{subfigure}{.24\textwidth}
  \centering
  \includegraphics[width=\linewidth]{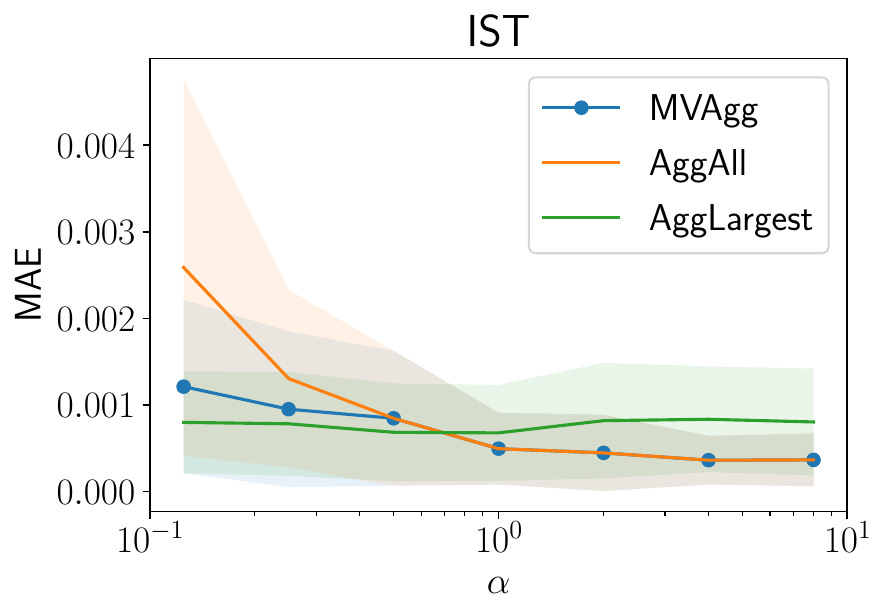}
\end{subfigure}
\begin{subfigure}{.24\textwidth}
  \centering
  \includegraphics[width=\linewidth]{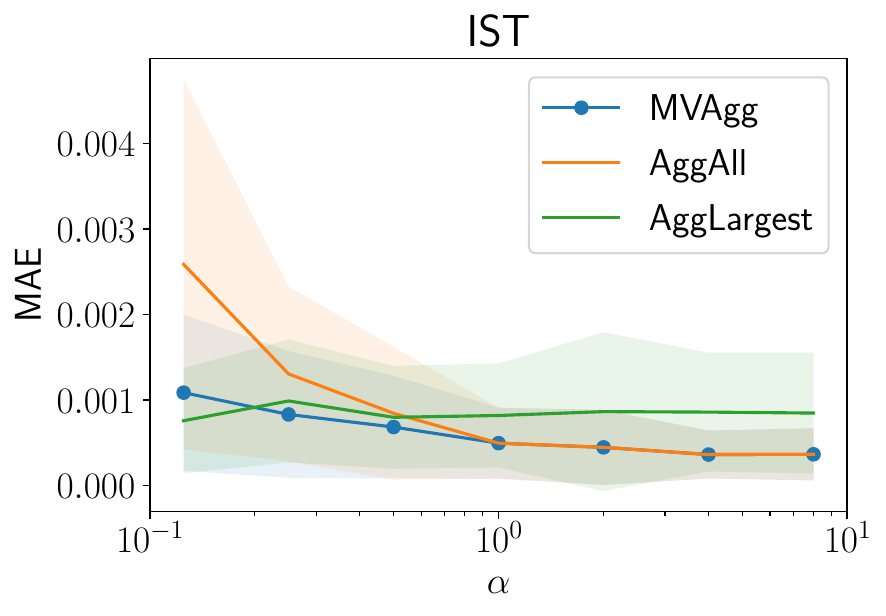}
\end{subfigure}
\begin{subfigure}{.24\textwidth}
  \centering
  \includegraphics[width=\linewidth]{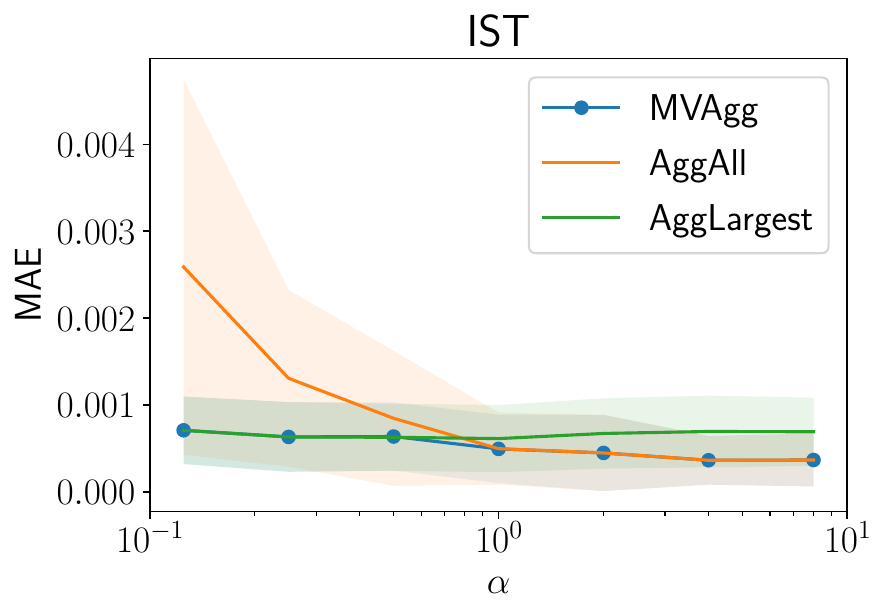}
\end{subfigure}
\caption{Mean MAEs and standard deviations of MVAgg, AggAll and AggLargest on IST dataset under three-site setting. $N_1:N_2:N_3 = 1:1:1$ (left most), $3:2:1$ (middle left), $9:9:2$ (middle right), and $18:1:1$ (right most).}
\label{fig:three-site-ist}
\end{figure}
Figure~\ref{fig:ist-two-varying-eps} shows the mean MAEs on IST dataset under the two-site setting ($J=2$). 
As $\alpha$ gets larger, the noise variance for the second site gets smaller while the one for the first site remains the same; thus, we generally expect the final ATE estimate to be never less accurate. We confirm this is true for all aggregation algorithms.
We see that MVAgg generally achieves the best MAE among the three aggregation methods. For most of $\alpha = \epsilon_2/\epsilon_1$, its MAE matches with the better one of AggAll and AggLargest. 
This suggests that when $\alpha$ is very small meaning the second site sends very noisy statistics, MVAgg discards the noisy site and only uses the results from the first site.
On the other hand, when $\epsilon_2$ is relatively large and the statistics from the second site are less noisy, MVAgg uses both sites to reduce a sampling error.
We also observe the standard deviations of MVAgg are mostly the smallest, which is because MVAgg aims to minimize the variance of ATE estimate.
AggAll performs the worst when $\epsilon_1\gg \epsilon_2$, which supports our intuition that the noisy site can harm the final ATE estimation.
The performance of AggLargest gets relatively worse as $\alpha$ gets larger since it does not utilize the accurate statistics from the second site.

Figure~\ref{fig:three-site-ist} shows the mean MAEs on IST dataset under the three-site setting ($J=3$) with varying sample size proportions. 
The results exhibit similar trends to the two-site one. Most notably, MVAgg outperforms AggAll and AggLargest in most of the cases.
Comparing the results of different sample size proportions, we see the performance gap between MVAgg and AggLargest is maximized when the sample distribution across sites is uniform, i.e., $N_1:N_2:N_3 = 1:1:1$. This is because AggLargest cannot use large enough sites even when those sites have large enough $\epsilon$'s.
On the other hand, the gap between MVAgg and AggAll for $\alpha \ll 1$ is largest when $N_1:N_2:N_3 = 18:1:1$. This is because AggAll weighs too much on the largest site, i.e., the first site, even when it has small $\epsilon_1$, leading to noisier results than the ones obtained by removing the first site. 
This case particularly demonstrates the non-triviality of the problem---more samples do not necessarily help the final ATE estimation in the presence of DP noise.

Figure~\ref{fig:star-four-varying-eps} shows the mean MAEs for STAR dataset, where the assignments to the four sites ($J=4$) are pre-determined. 
We observe similar trends for all three aggregation algorithms to the case on IST dataset.
Particularly on STAR dataset, MVAgg performs the best for all $\alpha$'s. This is because MVAgg has more flexibility to choose the number of sites used, e.g., it can use sites 1 to 3 while others cannot.

\subsubsection{Observational Study and Matching Estimator}
\begin{figure}[t]
    \centering
    \minipage{0.48\textwidth}
      \includegraphics[width=\linewidth]{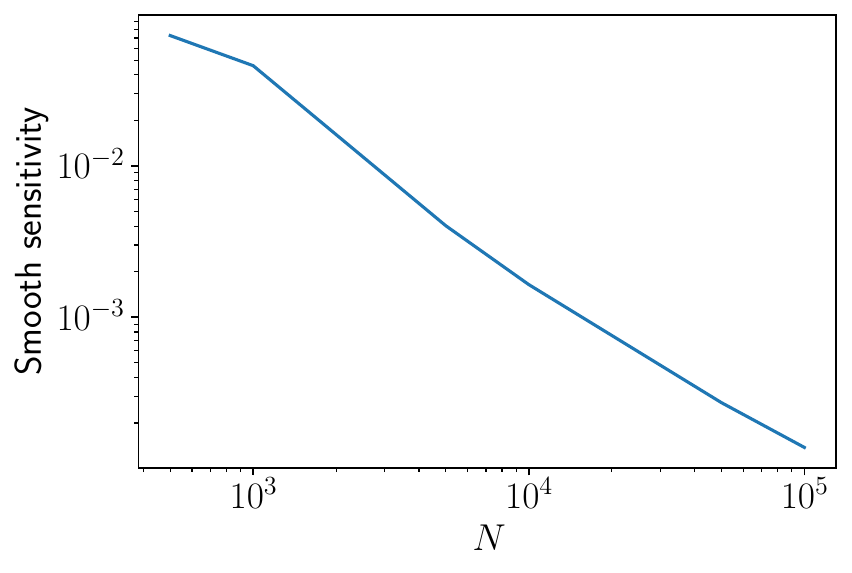}
      \caption{Smooth sensitivity of Synth dataset with varying sample sizes ($N$) under $|\mathcal{X}|=100$.}
      \label{fig:smooth-sens-by-Ns}
    \endminipage\hfill
    \minipage{0.48\textwidth}%
      \includegraphics[width=\linewidth]{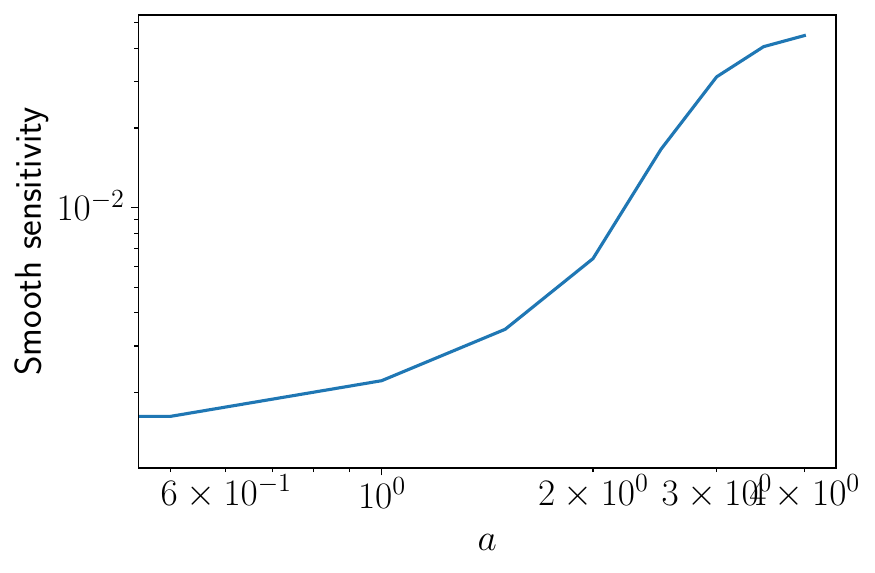}
      \caption{Smooth sensitivity of Synth dataset with varying extent of imbalance ($a$; larger $a$ yields more imbalanced dataset) under $N=10000$ and $|\mathcal{X}|=100$.}
      \label{fig:smooth-sens-by-as}
    \endminipage
\end{figure}
\begin{figure}[t]
\begin{subfigure}{.33\textwidth}
  \centering
  \includegraphics[width=\linewidth]{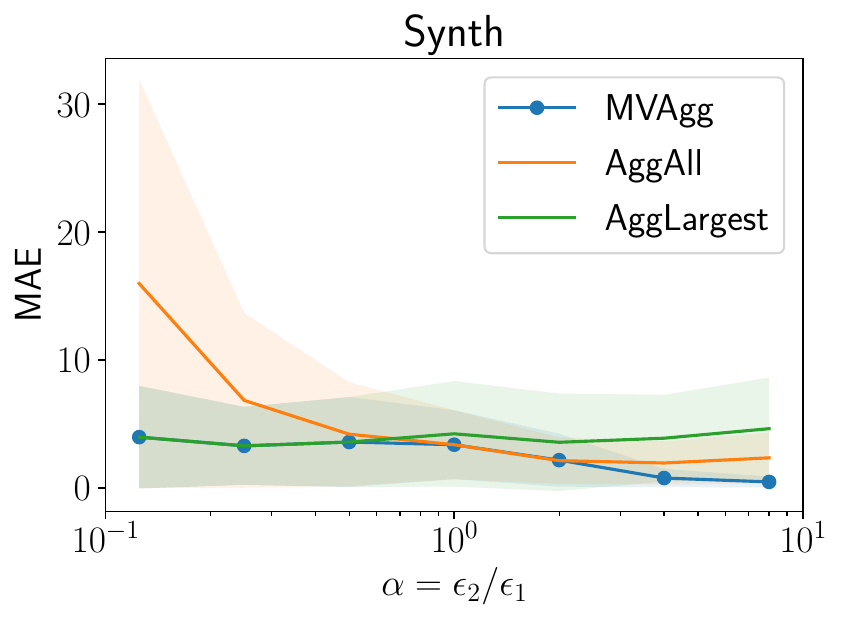}
\end{subfigure}%
\begin{subfigure}{.33\textwidth}
  \centering
  \includegraphics[width=\linewidth]{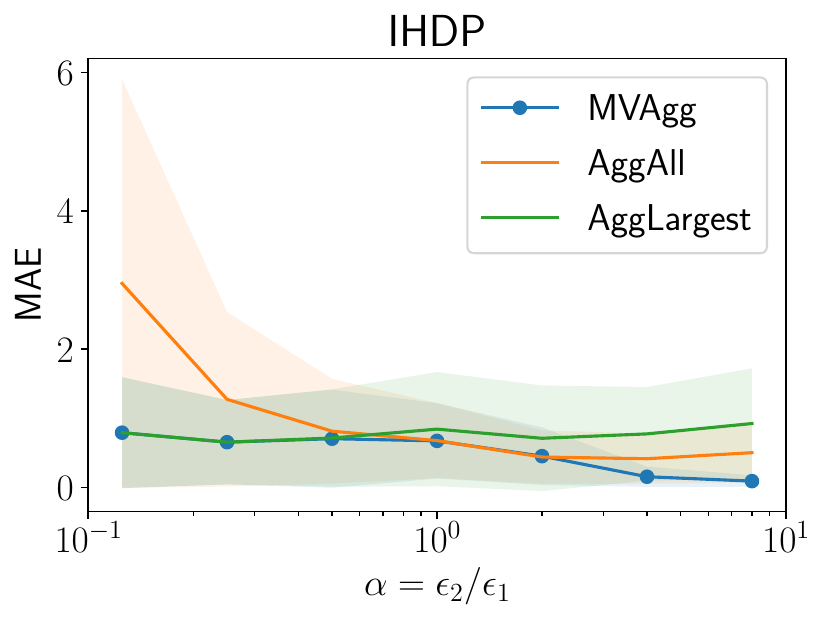}
\end{subfigure}
\begin{subfigure}{.33\textwidth}
  \centering
  \includegraphics[width=\linewidth]{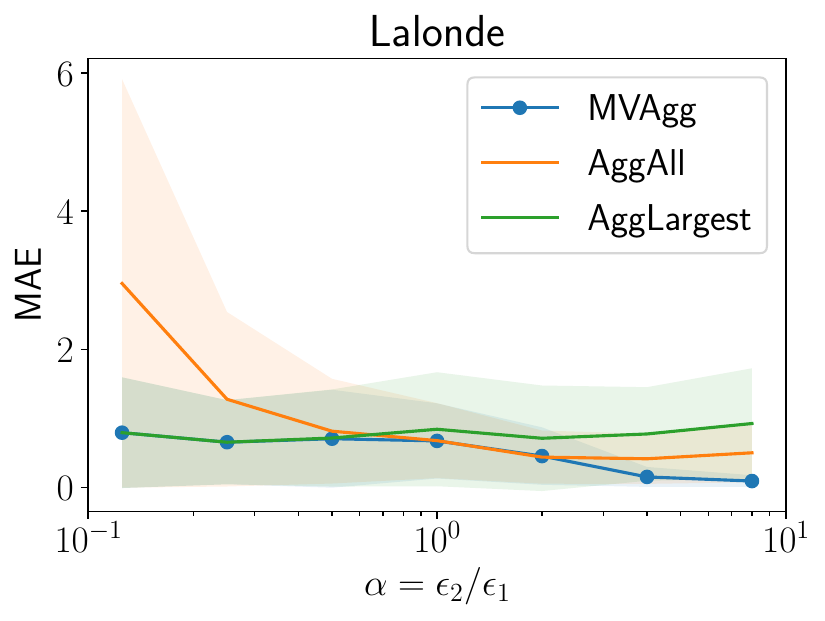}
\end{subfigure}
\begin{subfigure}{.33\textwidth}
  \centering
  \includegraphics[width=\linewidth]{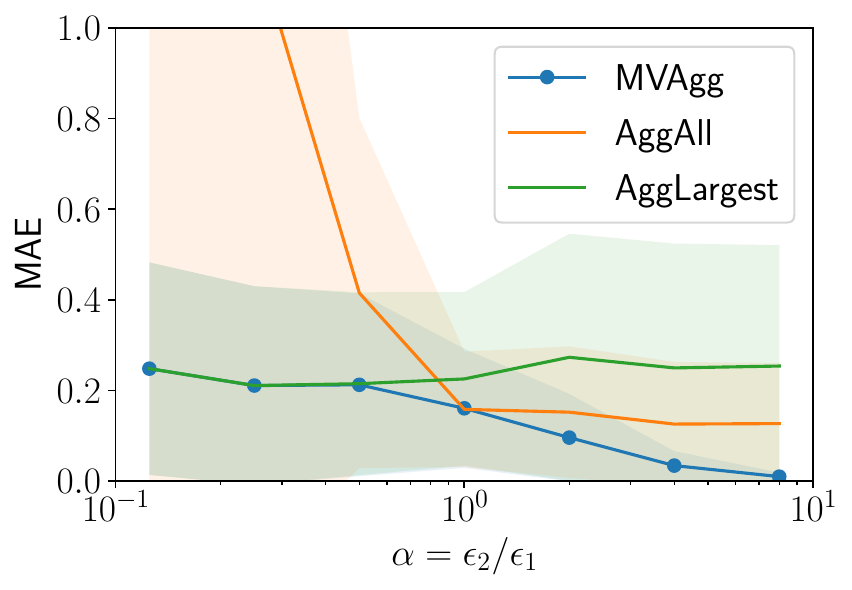}
\end{subfigure}%
\begin{subfigure}{.33\textwidth}
  \centering
  \includegraphics[width=\linewidth]{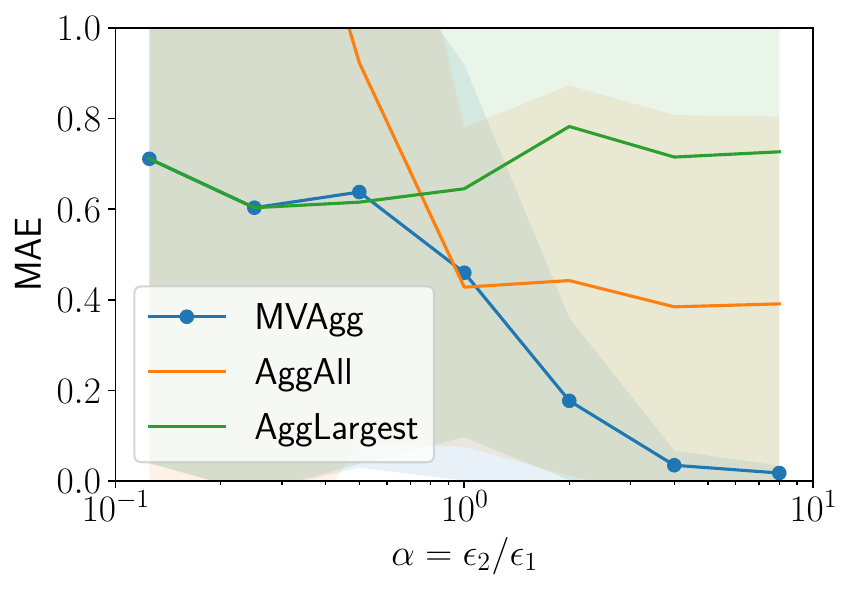}
\end{subfigure}
\begin{subfigure}{.33\textwidth}
  \centering
  \includegraphics[width=\linewidth]{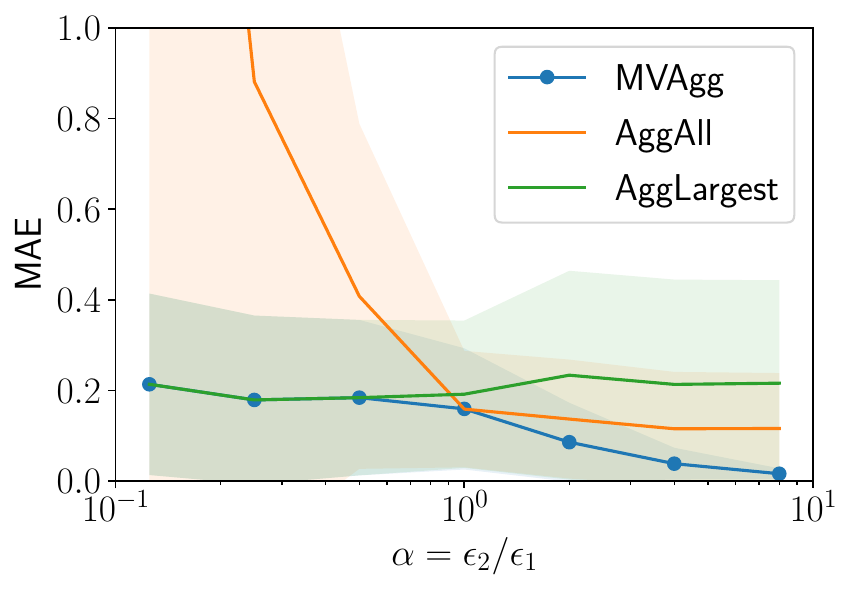}
\end{subfigure}
\caption{Mean MAEs and standard deviations of MVAgg, AggAll and AggLargest on Synth, IHDP, and Lalonde datasets (from left to right) under two-site setting. Upper row: GlobalDPMatching. Lower row: SmoothDPMatching. \textbf{Note that y-axis scales are different between upper and lower rows.}}
\label{fig:obs-varying-eps}
\end{figure}
\begin{figure}[t]
\centering
\begin{subfigure}{.24\textwidth}
  \centering
  \includegraphics[width=\linewidth]{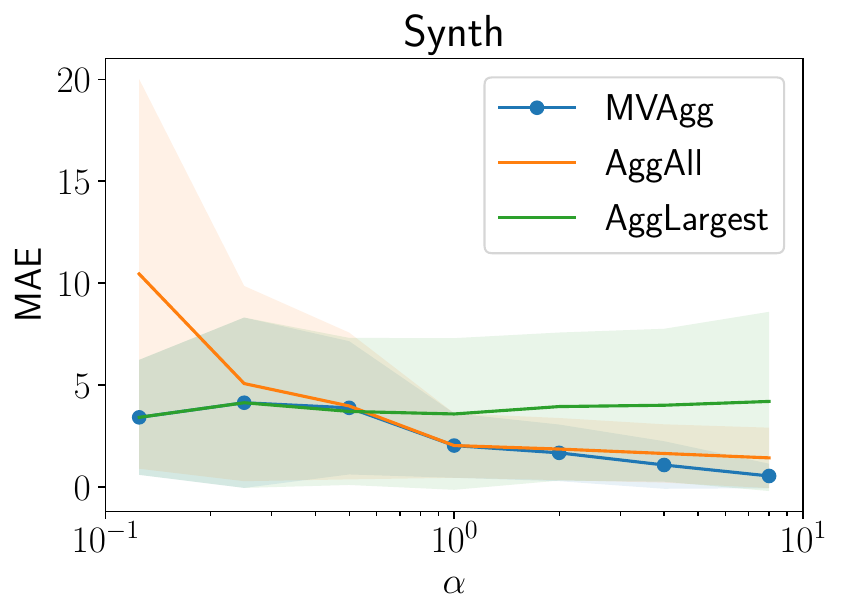}
\end{subfigure}%
\begin{subfigure}{.24\textwidth}
  \centering
  \includegraphics[width=\linewidth]{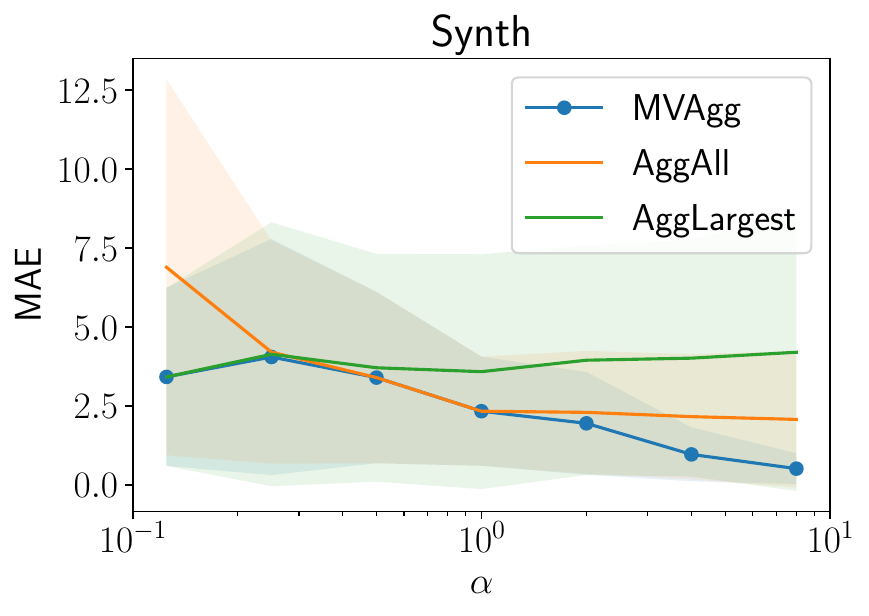}
\end{subfigure}
\begin{subfigure}{.24\textwidth}
  \centering
  \includegraphics[width=\linewidth]{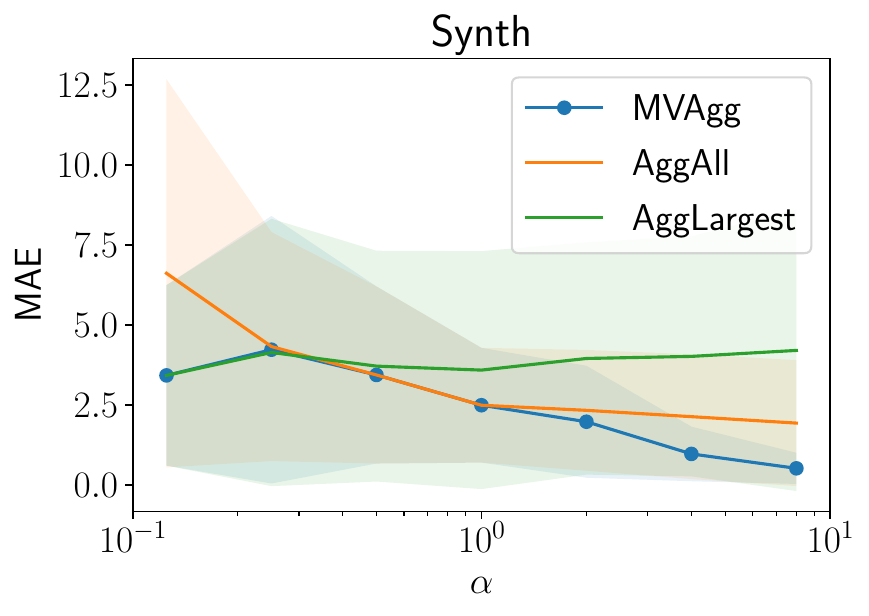}
\end{subfigure}
\begin{subfigure}{.24\textwidth}
  \centering
  \includegraphics[width=\linewidth]{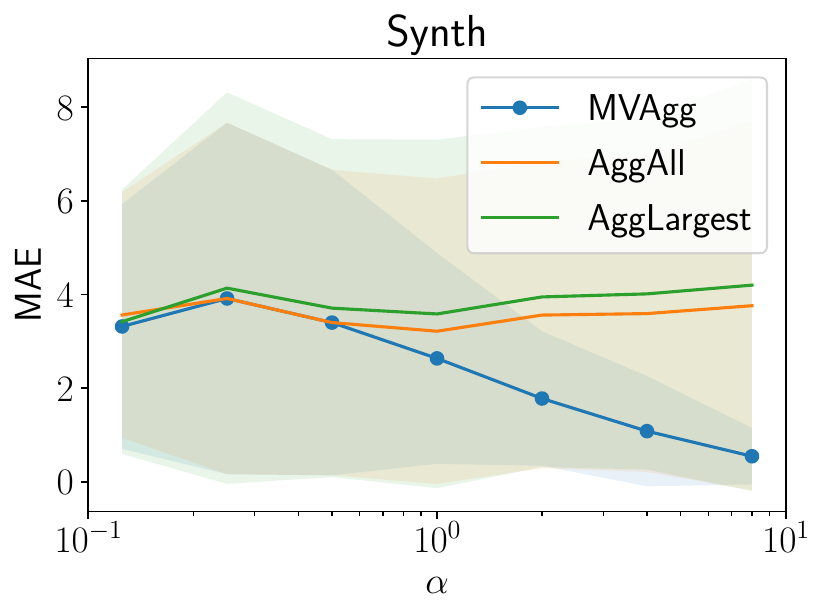}
\end{subfigure}
\begin{subfigure}{.24\textwidth}
  \centering
  \includegraphics[width=\linewidth]{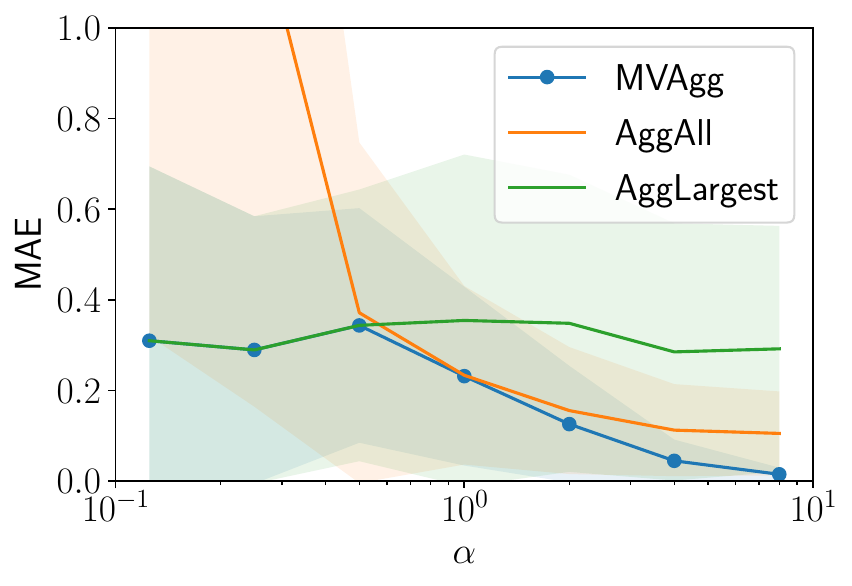}
\end{subfigure}%
\begin{subfigure}{.24\textwidth}
  \centering
  \includegraphics[width=\linewidth]{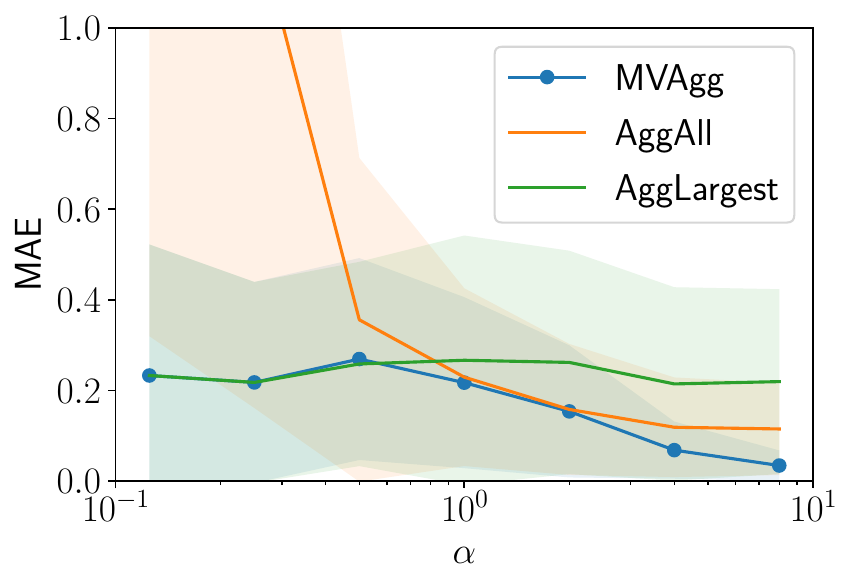}
\end{subfigure}
\begin{subfigure}{.24\textwidth}
  \centering
  \includegraphics[width=\linewidth]{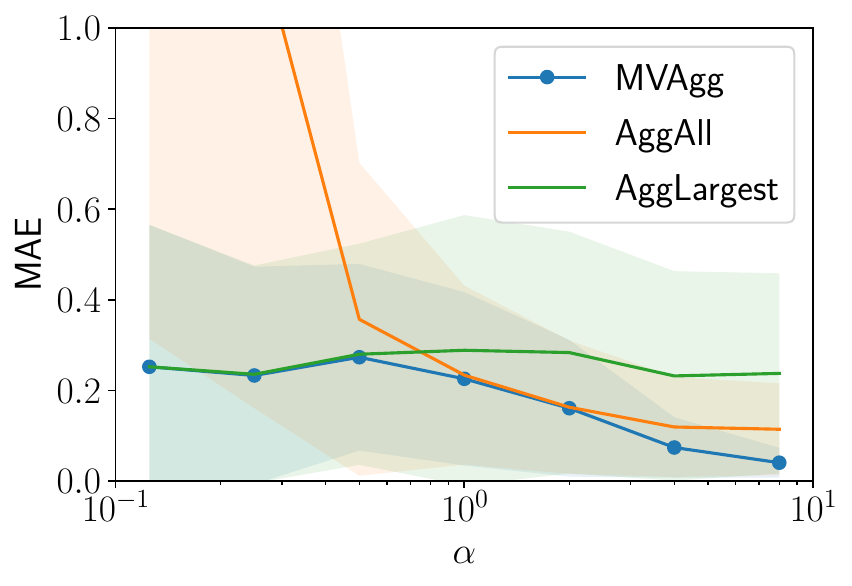}
\end{subfigure}
\begin{subfigure}{.24\textwidth}
  \centering
  \includegraphics[width=\linewidth]{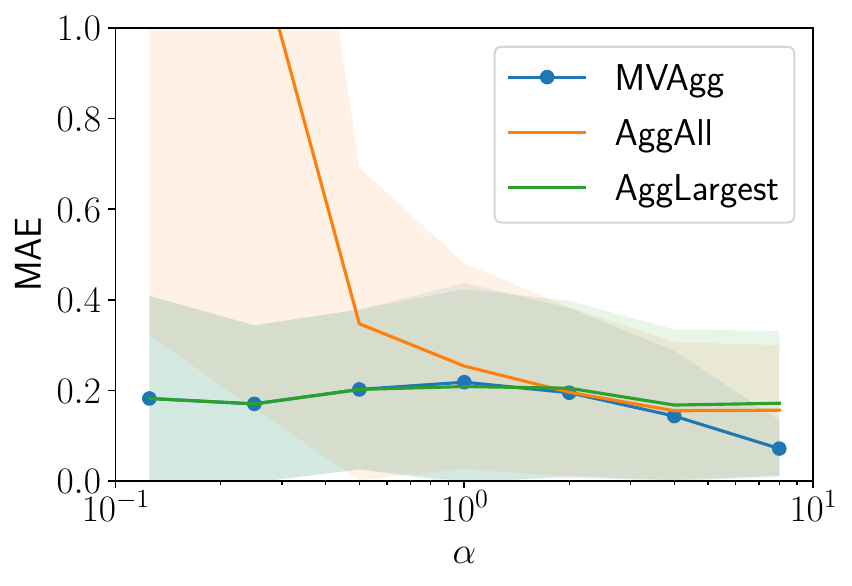}
\end{subfigure}
\caption{Mean MAEs and standard deviations of MVAgg, AggAll and AggLargest on Synth dataset under three-site setting. Upper low: GlobalDPMatching. Lower row: SmoothDPMatching. $N_1:N_2:N_3 = 1:1:1$ (left most), $3:2:1$ (middle left), $9:9:2$ (middle right), and $18:1:1$ (right most). \textbf{Note that y-axis scales are different between upper and lower rows.}}
\label{fig:three-site-obs-synth}
\end{figure}
\begin{figure}[t]
\centering
\begin{subfigure}{.24\textwidth}
  \centering
  \includegraphics[width=\linewidth]{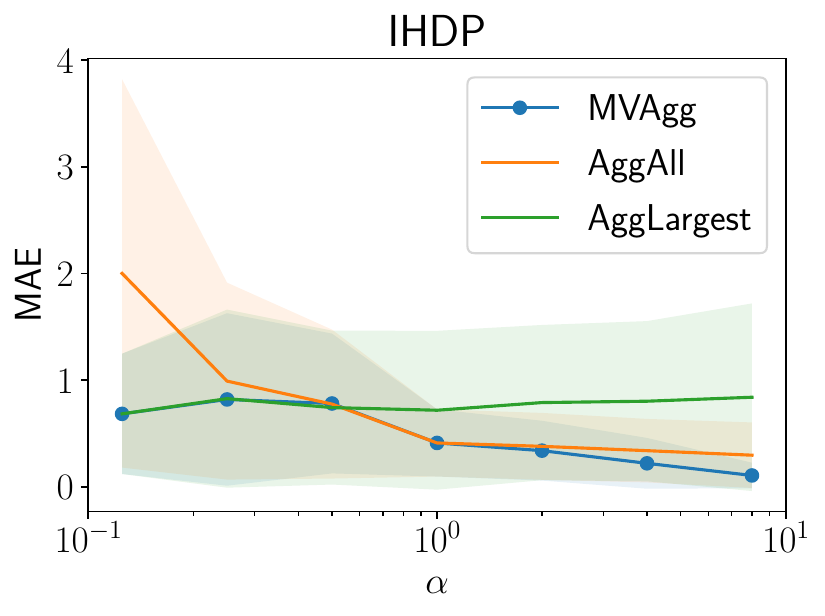}
\end{subfigure}%
\begin{subfigure}{.24\textwidth}
  \centering
  \includegraphics[width=\linewidth]{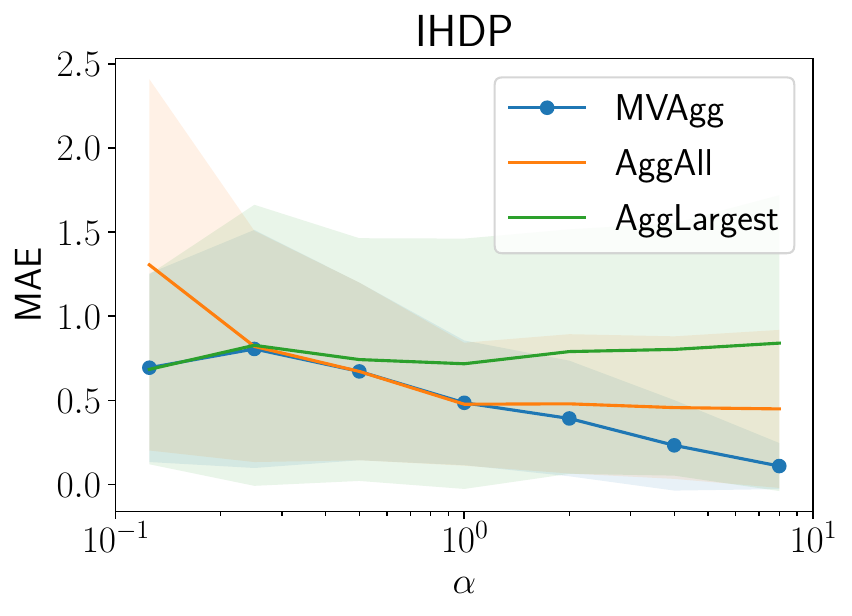}
\end{subfigure}
\begin{subfigure}{.24\textwidth}
  \centering
  \includegraphics[width=\linewidth]{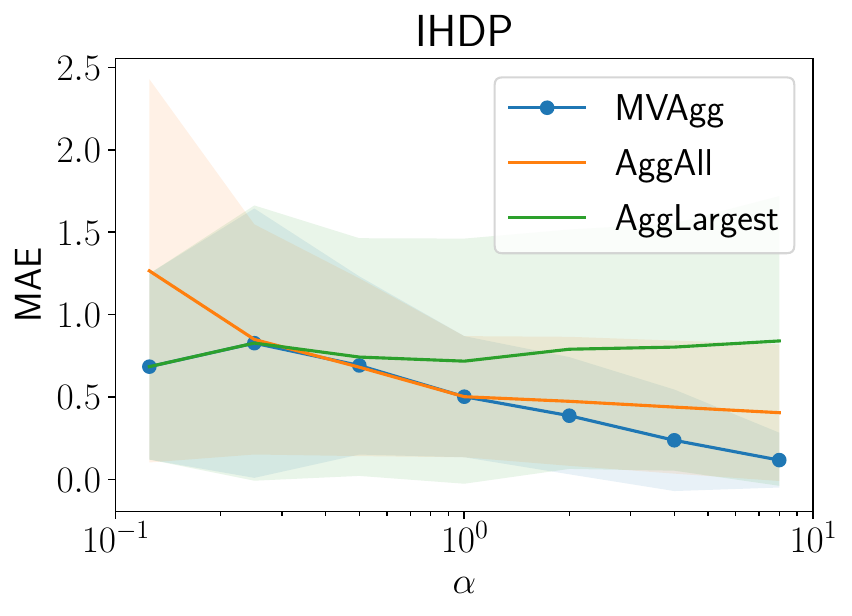}
\end{subfigure}
\begin{subfigure}{.24\textwidth}
  \centering
  \includegraphics[width=\linewidth]{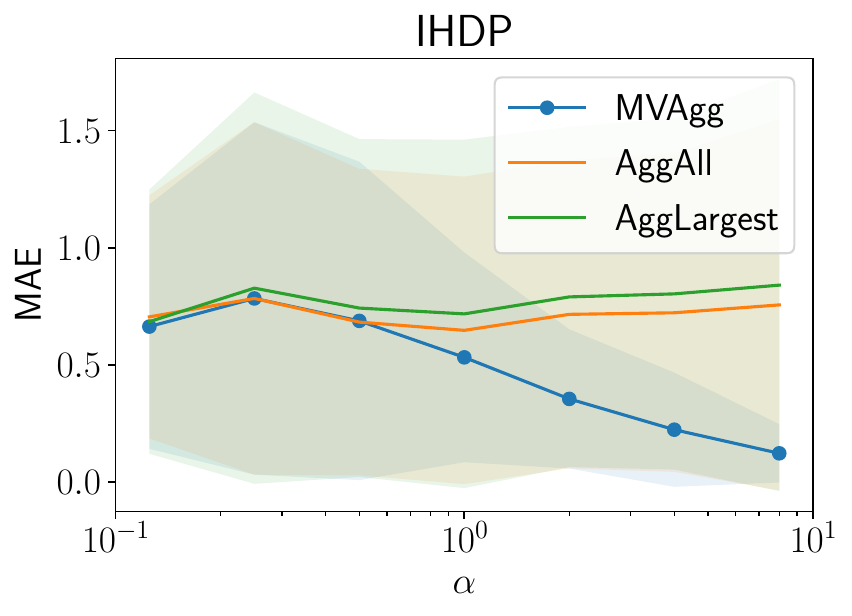}
\end{subfigure}
\begin{subfigure}{.24\textwidth}
  \centering
  \includegraphics[width=\linewidth]{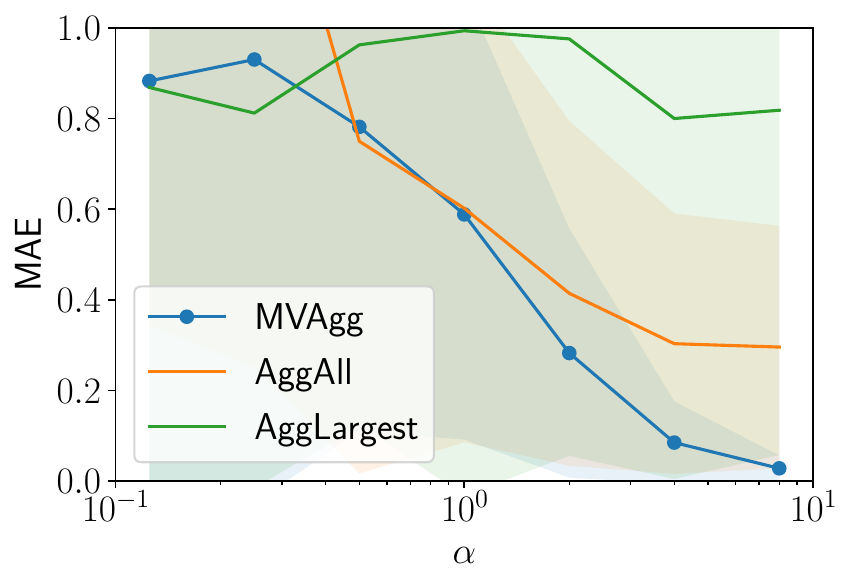}
\end{subfigure}%
\begin{subfigure}{.24\textwidth}
  \centering
  \includegraphics[width=\linewidth]{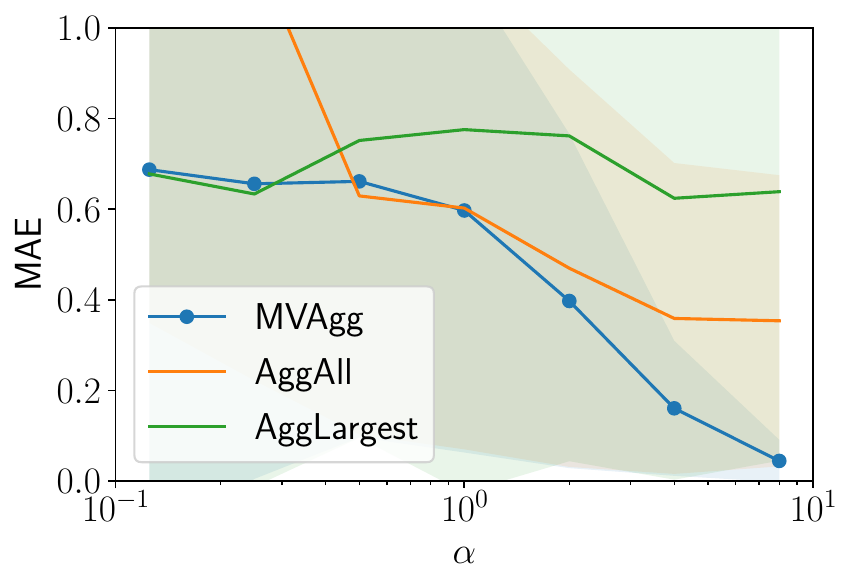}
\end{subfigure}
\begin{subfigure}{.24\textwidth}
  \centering
  \includegraphics[width=\linewidth]{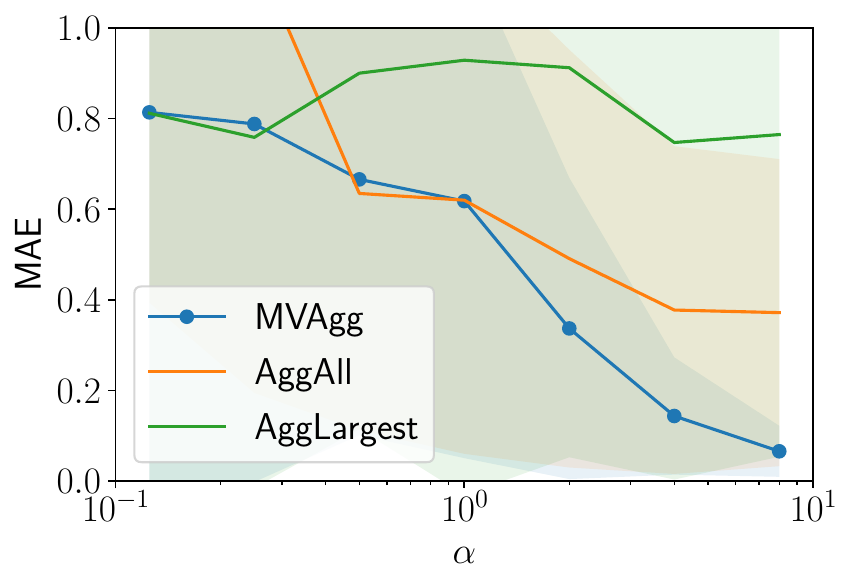}
\end{subfigure}
\begin{subfigure}{.24\textwidth}
  \centering
  \includegraphics[width=\linewidth]{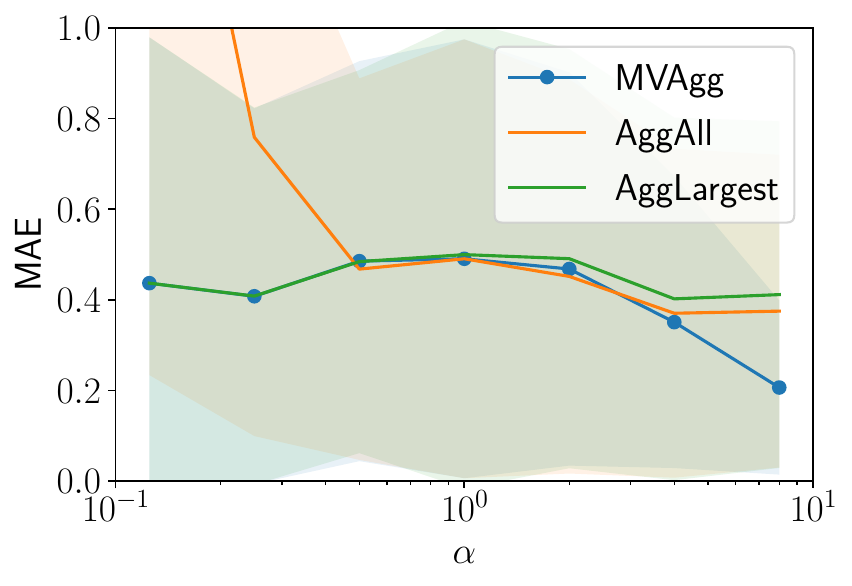}
\end{subfigure}
\caption{Mean MAEs and standard deviations of MVAgg, AggAll and AggLargest on IHDP dataset under three-site setting. Upper low: GlobalDPMatching. Lower row: SmoothDPMatching. $N_1:N_2:N_3 = 1:1:1$ (left most), $3:2:1$ (middle left), $9:9:2$ (middle right), and $18:1:1$ (right most). \textbf{Note that y-axis scales are different between upper and lower rows.}}
\label{fig:three-site-ihdp}
\end{figure}
\begin{figure}[t]
\centering
\begin{subfigure}{.24\textwidth}
  \centering
  \includegraphics[width=\linewidth]{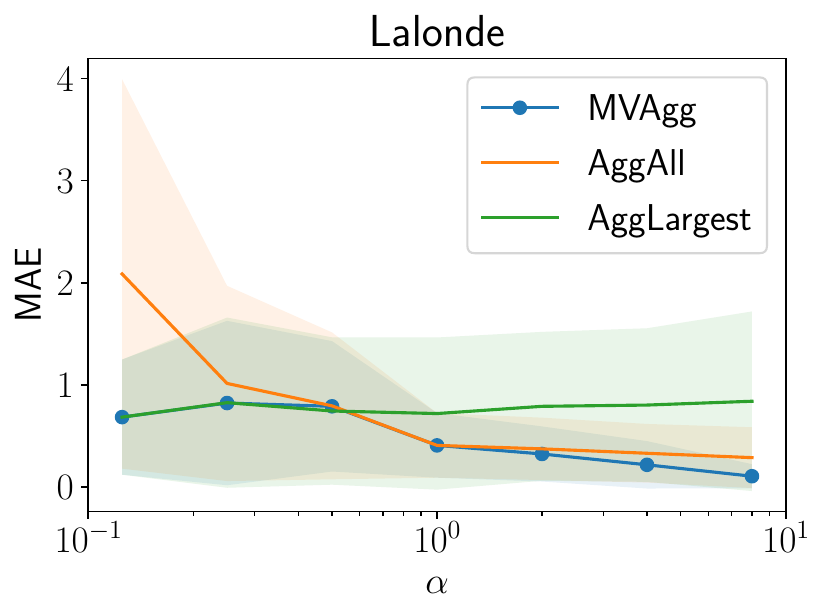}
\end{subfigure}%
\begin{subfigure}{.24\textwidth}
  \centering
  \includegraphics[width=\linewidth]{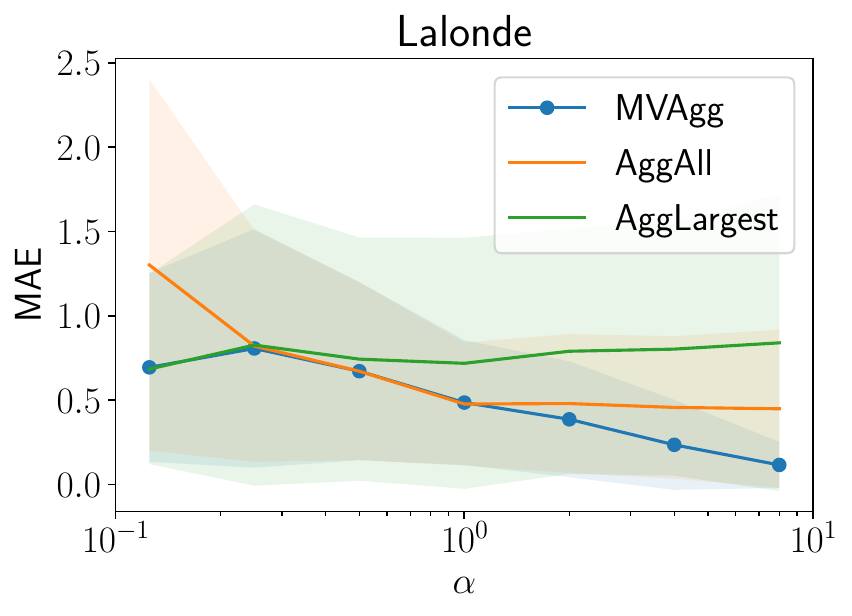}
\end{subfigure}
\begin{subfigure}{.24\textwidth}
  \centering
  \includegraphics[width=\linewidth]{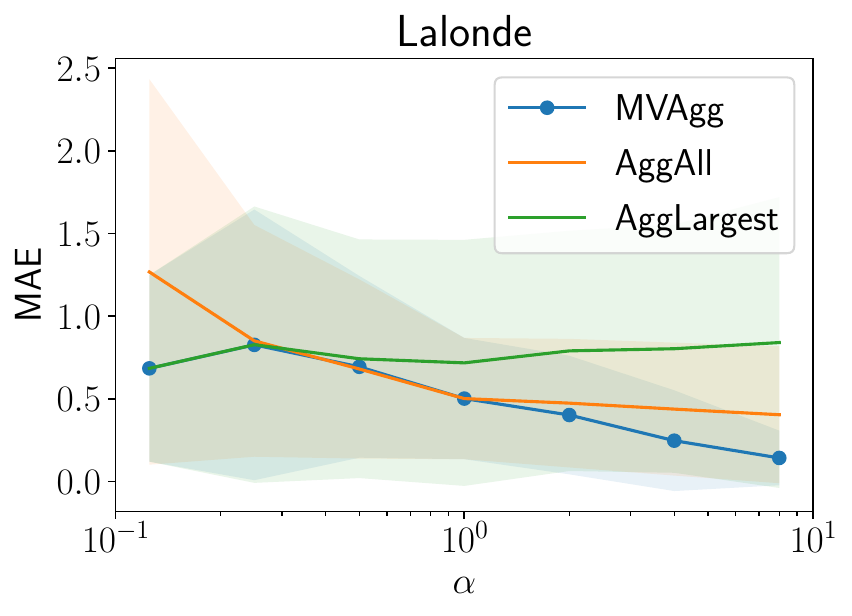}
\end{subfigure}
\begin{subfigure}{.24\textwidth}
  \centering
  \includegraphics[width=\linewidth]{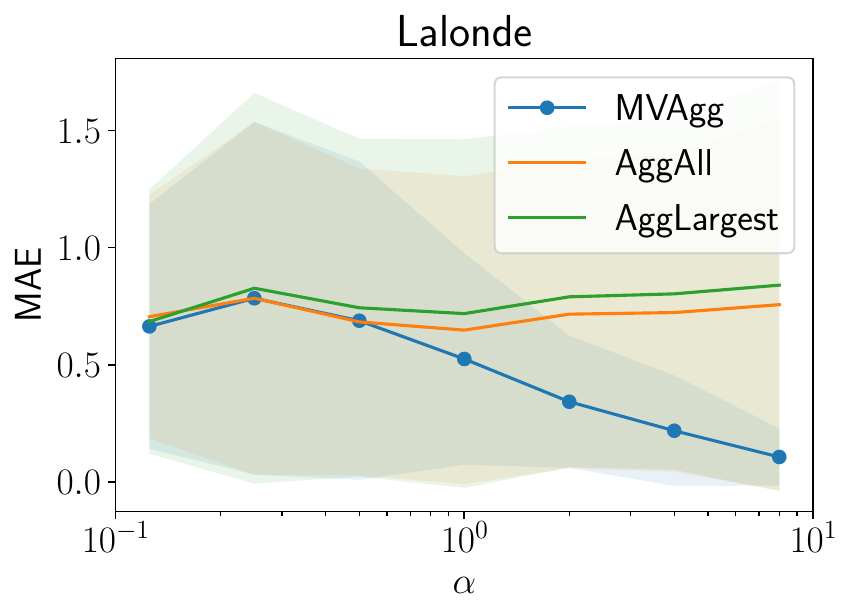}
\end{subfigure}
\begin{subfigure}{.24\textwidth}
  \centering
  \includegraphics[width=\linewidth]{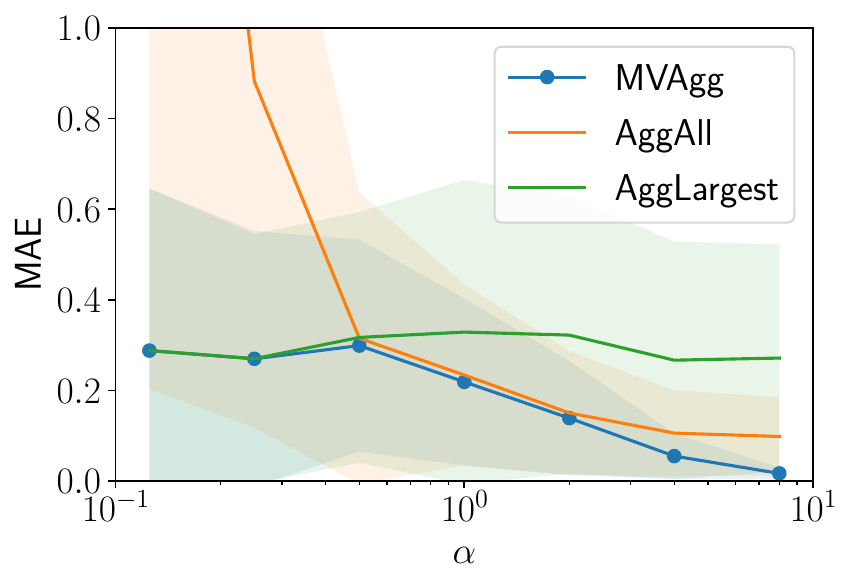}
\end{subfigure}%
\begin{subfigure}{.24\textwidth}
  \centering
  \includegraphics[width=\linewidth]{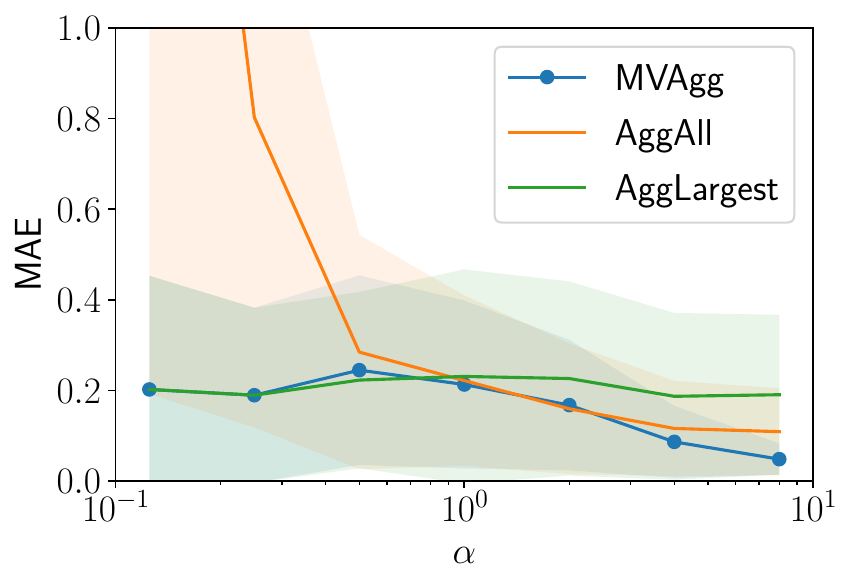}
\end{subfigure}
\begin{subfigure}{.24\textwidth}
  \centering
  \includegraphics[width=\linewidth]{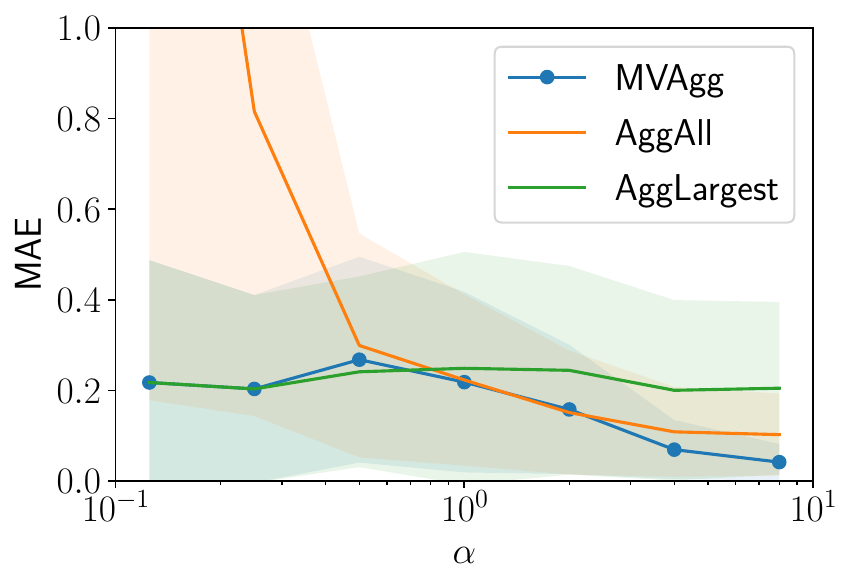}
\end{subfigure}
\begin{subfigure}{.24\textwidth}
  \centering
  \includegraphics[width=\linewidth]{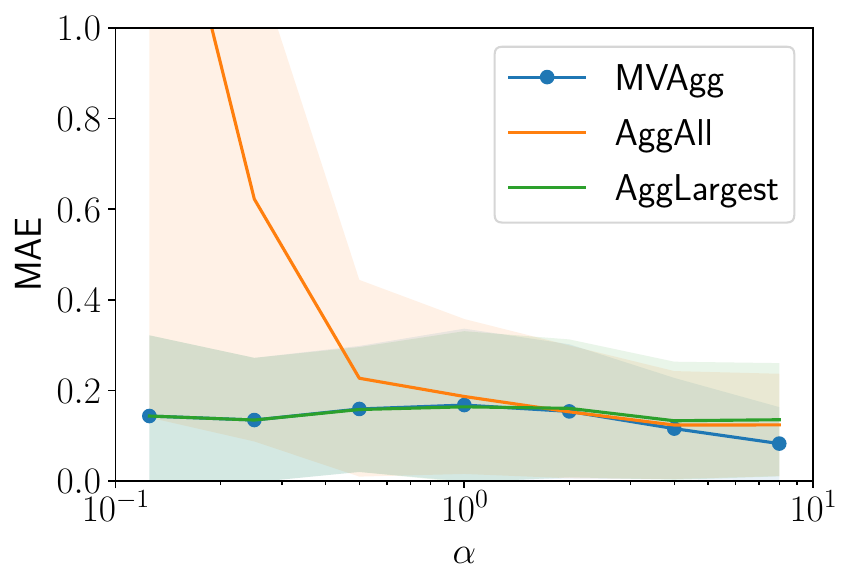}
\end{subfigure}
\caption{Mean MAEs and standard deviations of MVAgg, AggAll and AggLargest on Lalonde dataset under three-site setting. Upper low: GlobalDPMatching. Lower row: SmoothDPMatching. $N_1:N_2:N_3 = 1:1:1$ (left most), $3:2:1$ (middle left), $9:9:2$ (middle right), and $18:1:1$ (right most). \textbf{Note that y-axis scales are different between upper and lower rows.}}
\label{fig:three-site-lalonde}
\end{figure}
Figures~\ref{fig:smooth-sens-by-Ns} and~\ref{fig:smooth-sens-by-as} demonstrate how the smooth sensitivity on the Synth dataset changes along with the sample size $N$ and the extent of imbalance, which is controlled by the parameter $a$ (the larger $a$ is, the more imbalanced the dataset is).
Here, we measure $\beta$-smooth sensitivity for $\beta = \frac{\epsilon}{2\ln(\frac{2}{\delta})}$, where $\epsilon=1$ and $\delta=10^{-5}$, c.f., the smooth-sensitivity-based Laplace mechanism.
We observe that the smooth sensitivity actually scales with $\approx \mathcal{O}(1/N)$ for a balanced dataset.
We also see that it positively correlates with the extent of imbalance---it is the smallest when the data is well-balanced.
Notice that even for imbalanced data, the smooth sensitivity is smaller than the global sensitivity $\approx 1$.

Figures~\ref{fig:obs-varying-eps}--~\ref{fig:three-site-lalonde} show the MAEs on three observational study datasets, Synth, IHDP, and Lalonde, when the ATE estimation algorithm is GlobalDPMatching or SmoothDPMatching under two-site (Figure~\ref{fig:obs-varying-eps}) and three-site (Figure~\ref{fig:three-site-obs-synth}--~\ref{fig:three-site-lalonde}) settings.
Overall, we observe similar trends for all three aggregation algorithms to the case on the randomized trial datasets.
Especially, for both ATE estimation algorithms and for all datasets, we see MVAgg outperforms the other aggregation methods in general. 
Furthermore, we see the standard deviations of MVAgg are the smallest. In particular, when $\alpha \gg 1$, they are much smaller than the others. 
These trends suggest that MVAgg discards the sites with small $\epsilon$'s, where the additive noises dominate the site outputs, and only uses the sites with large enough $\epsilon$'s, where the noises are negligible, to reduce a sampling error. 

One main difference from the randomized trial case is the error scale. The global and smooth sensitivities of the matching estimator are larger than the global sensitivity of the difference-in-means estimator, which we use for randomized trials. 
Therefore, the ATE estimates at each site by the DP matching estimators, GlobalDPMatching and SmoothDPMatching, tend to be noisier, which results in higher MAEs.
In such a case, it is more beneficial to use MVAgg instead of AggAll and AggLargest since the absolute gains in MAE are much larger.  

Comparing the ATE estimation algorithms, we observe that SmoothDPMatching achieves much better performance (notice the scales of y-axis).
The MAEs of GlobalDPMatching can be around $1$ or more which is impermissible considering that $0\leq \tau \leq 1$ as a result of preprocessing.
However, SmoothDPMatching combined with MVAgg achieves MAEs less than $1$ for all cases and even achieves MAEs around $0.1$ or less when $\alpha \gg 1$.
This indicates that our smooth sensitivity analysis enables us to dramatically reduce an additive DP noise variance and improve the privacy-utility tradeoff.

\subsection{Discussion}

Our results support the superiority of SmoothDPMatching over GlobalDPMatching, which happens because the smooth sensitivity is much smaller than the global sensitivity in practice. 
We also anticipate that the advantage of SmoothDPMatching is larger for well-balanced datasets. 

Second, we find that as expected, MVAgg achieves the best final ATE on both randomized trial and observational study data, compared with the other rule-based aggregation algorithms. This is because it reliably adopts the estimate at a site only when the quality is relatively high. 
The relative quality is hugely dependent on the data through sampling error, which we cannot know in advance. Thus, MVAgg provides the principled way to aggregate the estimates from multiple sites as opposed to some other rule-based aggregation algorithm, e.g., AggAll and AggLargest.

Finally, we find that site-level privacy parameters also have a high impact on performance. In particular, when all sites have comparable privacy, it is best to combine their estimates; on the other hand, if some sites have significantly higher privacy requirements, then it is best not to use those sites. We find that MVAgg reliably does this for a variety of privacy parameters. 
Furthermore, we note that MVAgg never outputs the impermissible outcome for any combination of privacy parameters across sites. Considering the risk of outputting very noisy final estimates with rule-based algorithms, it is recommended to use MVAgg in general.

\section{Conclusion and Future Work}
We introduce a multi-site ATE estimation setting with per-site DP guarantees. We then provide a class of per-site ATE estimation algorithms which output both the private ATE estimate and its private variance estimate so that the central server aggregates the estimates from sites properly by looking at their qualities. In particular, for observational study data, we propose a novel DP matching estimator by analyzing the smooth sensitivity. We also propose an aggregation algorithm on the server that minimizes the variance of the final ATE estimate.
Our experimental results demonstrate that our method, combining our site and server algorithms, automatically handles the heterogeneity across sites and provides a better privacy-utility tradeoff.

We believe our work is a first step towards enabling causal inference studies across multiple sites with formal privacy guarantees. One of the future directions is to consider how we can combine statistics from sites with different data distributions, e.g., children's hospitals and geriatric hospitals. Another direction would be studying other estimands, e.g., CATE, and other estimators, e.g., IPW.

\section*{Acknowledgments}
TK and KC would like to thank 
NSF under 1804829,
NSF under 2241100,
NSF under 2217058,
ARO MURI W911NF2110317, and
ONR under N00014-20-1-2334
for research support. 
Also, TK is supported in part by Funai Overseas Fellowship.

\bibliographystyle{abbrv}
\bibliography{ms}

\end{document}


\appendix

\section{Detail on Modification of Matching Estimator}
We modify the exact single matching estimator so as to balance the number of individuals matched to a particular individual in the same covariate stratum.
The reason for doing this is to obtain the tighter bound of the sensitivity.
In particular, suppose for a stratum $X=x$, we have $|T_x|$ treated and $|C_x|$ control individuals. Without specifying any, the first control individual can be matched with all $|T_x|$ treated in the worst case (left side of Figure 2 in the main paper). Thus, the first one contributes $|T_x|+1$ times to the final estimate which leads to larger sensitivity and more DP noise vaiance.
As such, we avoid such cases by greedily balancing the number of matches for each individual (right side of Figure 2 in the main paper) within each covariate stratum. More specifically, $i$-th treated individual is matched with $i\bmod |C_x|$-th control, and $j$-th control individual is matched with $j \bmod |T_x|$-th treated. This way we guarantee that each treated (or control) individual contributes up to $\lceil |C_x|/|T_x|\rceil + 1$ (or $\lceil |T_x|/|C_x|\rceil + 1$) times to the final estiamte. As a result, we obtain the tighter sensitivity bound.
Note that we only specify how to handle individuals with the same covariate in the exact single matching; thus, no bias is introduced by our modification.

\section{Proof of Theorem 1} \label{sec:proof-thm1}
\begin{proof}
Let $L_i$ be the number of individuals matched with $i$-th individual, namely, the number of individuals who use $Y_i^\mathrm{obs}$ as the imputed value.
By our modification to the estimator in the main paper, we have $\lfloor\frac{|C_x|}{|T_x|}\rfloor \leq L_i \leq \lceil\frac{|C_x|}{|T_x|}\rceil$ when $X_i=x$ and $W_i=1$.
Then, the exact single matching estimator is written as below.
\begin{align*}
    \hat{\tau}(D) = \frac{1}{N}\sum_x \sum_{i\in T_x} (1 + L_i) Y_i^\mathrm{obs} - \sum_{i\in C_x} (1 + L_i) Y_i^\mathrm{obs}
\end{align*}
Let $f(D) = N\cdot\hat{\tau}(D)$. Since $\mathrm{LS}_{\hat{\tau}}(D) = \frac{1}{N}\mathrm{LS}_f(D)$, we instead consider the local sensitivity of $f$.

Additionally, let $d_{ar}(D,D^\prime)$ be the minimum number of row addition/removal from $D$ to obtain $D^\prime$. Furthermore, let $d_{ar}^+(D,D^\prime)\leq 1 \Leftrightarrow d_{ar}(D,D^\prime)\leq 1 \land |D^\prime|=|D|+1$ and define $\mathrm{LS}_f^{+}(D) = \max_{D^\prime: d_{ar}^+(D,D^\prime)\leq 1} |f(D)-f(D^\prime)|$.
Similarly, let $d_{ar}^-(D,D^\prime)\leq 1 \Leftrightarrow d_{ar}(D,D^\prime)\leq 1 \land |D^\prime|=|D|-1$ and define $\mathrm{LS}_f^{-}(D) = \max_{D^\prime:  d_{ar}^-(D,D^\prime)\leq 1} |f(D)-f(D^\prime)|$.
Then, by the triangle inequality, it holds that 
\begin{align*}
    \mathrm{LS}_f(D) &\leq
    \max_{D^{\prime\prime}: d_{ar}^+(D,D^{\prime\prime})\leq 1}\max_{D^\prime: d_{ar}^-(D^{\prime\prime},D^\prime)\leq 1}|f(D)-f(D^{\prime\prime})| + |f(D^{\prime\prime})-f(D^\prime)|\\
    &=\max_{D^{\prime\prime}: d_{ar}^+(D,D^{\prime\prime})\leq 1}(|f(D)-f(D^{\prime\prime})| +  
    \max_{D^\prime: d_{ar}^-(D^{\prime\prime},D^\prime)\leq 1}|f(D^{\prime\prime})-f(D^\prime)|)\\
    &\leq \mathrm{LS}^{+}_f(D) + \max_{D^{\prime\prime}: d_{ar}^+(D,D^{\prime\prime})\leq 1} \mathrm{LS}^{-}_f(D^{\prime\prime})
\end{align*} 

We first consider $\mathrm{LS}^{+}_f(D)$.
We write the neighboring dataset $D^\prime = \{(W_i^\prime, Y_i^{\mathrm{obs}\prime}, X_i^\prime)\}_{i=1}^{N+1}$ and also define $T^\prime_x$, $C^\prime_x$, and $L^\prime_i$ accordingly.
W.l.o.g., $D_{N+1}\in D^\prime$ is the added individual data.
Let $x$ be $x=X^\prime_{N+1}$. 

Here we assume w.l.o.g. $W_{N+1} = 1$. When $C_x,T_x$ are non-empty, it holds that
\begin{align*}
    |f(D)-f(D^\prime)| & 
    = |(\sum_{i\in T_x} (1 + L_i) Y_i^\mathrm{obs} - \sum_{i\in C_x} (1 + L_i) Y_i^\mathrm{obs}) 
    - (\sum_{i\in T_x^\prime} (1 + L_{i}^\prime) Y_i^{\mathrm{obs}\prime} - \sum_{i\in C_x^\prime} (1 + L_{i}^\prime) Y^{\mathrm{obs}\prime}_i)|\\
    & = |-(1 + L_{N+1}^\prime) Y^{\mathrm{obs}\prime}_{N+1} + \sum_{i\in T_x} (L_i - L_{i}^\prime) Y_i^\mathrm{obs}
    - \sum_{i\in C_x} (L_i - L_{i}^\prime) Y_i^\mathrm{obs}|\\
    &\leq |(1 + L_{N+1}^\prime) Y^{\mathrm{obs}\prime}_{N+1}| + |\sum_{i\in T_x} (L_i - L_{i}^\prime) Y_i^\mathrm{obs}|
    + |\sum_{i\in C_x} (L_i - L_{i}^\prime) Y_i^\mathrm{obs}|\\
    &\leq (1+ L_{N+1}^\prime) B + L_{N+1}^\prime B + B\\
    &= 2(1 + L_{N+1}^\prime)B\\
    &\leq 2(1+\lceil \frac{|C_x^\prime|}{|T_x^\prime|} \rceil)B
    = 2(1+\lceil \frac{|C_x|}{|T_x|+1} \rceil)B.
\end{align*}
The second to last inequality holds due to properties achieved by the (greedy) exact single matching estimator: (1)  for $i \in T_x$, $L_i \geq L_{i}^\prime$, (2)$\sum_{i\in T_x} L_i - L_{i}^\prime = L_{N+1}^\prime$ since $\sum_{i\in T_x} L_i = \sum_{i\in T_x^\prime} L_i^\prime = |C_x|$, and (3) for $i\in C_x$, there exists only one $i^\prime \in C_x$ s.t. $L_{i^\prime} - L^\prime_{i^\prime} = -1$ and for $i\neq i^\prime$, $L_i = L_{i}^\prime$.
In particular, the second term is bounded as follows.
\begin{align*}
    |\sum_{i\in T_x} (L_i - L_{i}^\prime) Y_i^\mathrm{obs}| \leq \sum_{i\in T_x} |L_i - L_{i}^\prime| |Y_i^\mathrm{obs}|
\leq B\sum_{i\in T_x} |L_i - L_{i}^\prime|
= B\sum_{i\in T_x} (L_i - L_{i}^\prime) = BL_{N+1}^\prime 
\end{align*}
The last inequality holds the exact single matching estimator balances the number of matches.

When $C_x$ is empty, $|f(D)-f(D^\prime)| = 0$.
When $T_x$ is empty and $C_x$ is not empty, $|f(D)-f(D^\prime)| = |Y_{N+1}^\mathrm{obs} - \hat{Y}_{N+1}(0) - \sum_{i\in C_x} Y_{N+1}^\mathrm{obs} - Y_i^\mathrm{obs}| \leq 2(|C_x|+1)B$.

Therefore, by the symmetry, the following holds.
\begin{align*}
   \mathrm{LS}^{+}_f(D) = \max_x \begin{cases} 0 & |T_x| = |C_x| = 0\\
   2(1+\max(\lceil \frac{|C_x|}{|T_x| + 1}\rceil, \lceil \frac{|T_x|}{|C_x| + 1}\rceil))B & o.w. 
   \end{cases} 
\end{align*}
With similar arguments, we have the following.
\begin{align*}
   \mathrm{LS}^{-}_f(D) = \max_x \begin{cases} 0 & |T_x| = 0 \lor |C_x| = 0\\
   2(1+\max(\lceil \frac{|C_x|}{|T_x|}\rceil, \lceil \frac{|T_x|}{|C_x|}\rceil))B & o.w.
   \end{cases} 
\end{align*}
Therefore, 
\begin{align*}
    \max_{D^{\prime\prime}: d_{ar}^+(D,D^{\prime\prime})\leq 1} \mathrm{LS}^{-}_f(D^{\prime\prime}) = 
    \max_x \begin{cases} 0 & |T_x| = |C_x| = 0\\
    2(1+|C_x|)B & |T_x| = 0 \land |C_x| > 0 \\
    2(1+|T_x|)B & |T_x| > 0 \land |C_x| = 0 \\
    2(1+\max(\lceil \frac{1+|C_x|}{|T_x|}\rceil, \lceil \frac{1+|T_x|}{|C_x|}\rceil))B & o.w.
    \end{cases}
\end{align*}

Finally, by combining above, we have the upper bound on the local sensitivity as follows.
\begin{align*}
    \mathrm{LS}_{\hat{\tau}}(D) &\leq \frac{1}{N}\max_x
    \begin{cases}
    0 & |T_x| = |C_x| = 0\\
    4(1+|C_x|)B & |T_x| = 0 \land |C_x| > 0 \\
    4(1+|T_x|)B & |T_x| > 0 \land |C_x| = 0 \\
    2(2+\max(\lceil \frac{|C_x|}{|T_x| + 1}\rceil, \lceil \frac{|T_x|}{|C_x| + 1}\rceil)+\max(\lceil \frac{1+|C_x|}{|T_x|}\rceil, \lceil \frac{1+|T_x|}{|C_x|}\rceil))B & o.w.
    \end{cases} \\
    &=\frac{1}{N}\max_x
    \begin{cases}
    0 & |T_x| = |C_x| = 0\\
    4(1+|C_x|)B & |T_x| = 0 \land |C_x| > 0 \\
    4(1+|T_x|)B & |T_x| > 0 \land |C_x| = 0 \\
    2(2+\max(\lceil \frac{|C_x|}{|T_x| + 1}\rceil + \lceil \frac{1+|C_x|}{|T_x|}\rceil, \lceil \frac{|T_x|}{|C_x| + 1}\rceil+\lceil \frac{1+|T_x|}{|C_x|}\rceil))B & o.w.
    \end{cases}\\
    &\leq\frac{1}{N}\max_x
    \begin{cases}
    0 & |T_x| = |C_x| = 0\\
    4(1+|C_x|)B & |T_x| = 0 \land |C_x| > 0 \\
    4(1+|T_x|)B & |T_x| > 0 \land |C_x| = 0 \\
    4(1+\max(\lceil \frac{1+|C_x|}{|T_x|}\rceil, \lceil \frac{1+|T_x|}{|C_x|}\rceil))B & o.w.
    \end{cases}
\end{align*}

The local sensitivity depends only on $|T_x|$ and $|C_x|$, and thus, the smooth sensitivity is obtained as follows.
Let $R_x(D)$ satisfy $\mathrm{LS}(D) = \frac{4B}{N}(1+\max_l R_x(D))$.
Also, let $R^{(k)}_x(D) = \max_{D^\prime: d(D,D^\prime)\leq k} R_x(D^\prime)$.
Then,
\begin{align*}
    A^{(k)}(D) &= \max_{D^\prime: d(D,D^\prime)\leq k} \mathrm{LS}(D^\prime)\\
    &= \frac{4B}{N} (1 + \max_x R^{(k)}_x(D)).
\end{align*}
Here, $R^{(k)}_x(D)$ is as follows.
\begin{align*}
    R_x{(k)} = \begin{cases}
    |T_x|+k & |T_x| \geq |C_x| \land k \geq |C_x|\\
    \lceil \frac{|T_x|+k+1}{|C_x|-k} \rceil &|T_x| \geq |C_x| \land k < |C_x|\\
    |C_x|+k & |C_x| \geq |T_x| \land k \geq |T_x|\\
    \lceil \frac{|C_x|+k+1}{|T_x|-k} \rceil &|C_x| \geq |T_x| \land k < |T_x|
    \end{cases}
\end{align*}

Thus, the $\beta$-smooth sensitivity is 
\begin{align*}
    S^*_{\hat{\tau},\beta}(D) = \max_{k=0,\ldots, N} e^{-k\beta} \frac{4B}{N}(1+\max_x R^{(k)}_x(D)).
\end{align*}
This can be computed by storing $|T_x|$ and $|C_x|$ for each $x\in\mathcal{X}$, which is present in the dataset, and enumerating over $k$; thus,
we need $\mathcal{O}(\min(|\mathcal{X}|,N))$ space and $\mathcal{O}(N\cdot \min(|\mathcal{X}|,N))$ time.
Note that for $x\in\mathcal{X}$ which is not present in th dataset $R_x^{(k)}(D)$ is constant for fixed $k$, i.e., $R_x^{(k)}(D)=k$.
\end{proof}

\section{Differentially Private Variance Estimation}
Let $\hat{\tau}_\mathrm{DP}$ be our DP matching estimator, SmoothDPMatching.
Recall that the variance of $\hat{\tau}_\mathrm{DP}$ is $\mathbb{V}[\hat{\tau}_{\mathrm{DP}}] = \mathbb{V}[\hat{\tau}] + \mathbb{V}[(2S^*_{\hat{\tau}, \beta}(D)/\epsilon) \cdot \eta] = \mathbb{V}[\hat{\tau}] + 8(S^*_{\hat{\tau},\beta}(D))^2/\epsilon^2$.
Since both terms are data-dependant, we need to estimate them from data privately.

We guarantee $(\epsilon_2,\delta_2)$-DP and $(\epsilon_3,\delta_3)$-DP separately for each term. Suppose the private ATE estimation is done with $(\epsilon_1,\delta_1)$-DP. Then, publishing the private ATE estimate and the variance estimate satisfies $(\epsilon_1+\epsilon_2+\epsilon_3,\delta_1+\delta_2+\delta_3)$-DP in total.

\subsection{Smooth Sensitivity of Variance of Matching Estimator}
By Section 19 of \cite{imbens_causal_2015}, we have the following variance estimate of the exact single matching estimator, $\hat{\mathbb{V}}[\hat{\tau}]$:
\begin{align*}
    \hat{\mathbb{V}}[\hat{\tau}] = \frac{1}{2N^2} \sum_x \left\{ \sum_{i\in T_x} (1+L_i)^2 (\hat{Y}_i(1)-\hat{Y}_i(0))^2 + \sum_{i\in C_x} (1+L_i)^2 (\hat{Y}_i(1)-\hat{Y}_i(0))^2\right\}.
\end{align*}
We produce DP version of the variance estimate by adding noise calibrated to the smooth sensitivity of this quantity. In the rest of this section, we present its smooth sensitivity.

As in Section~\ref{sec:proof-thm1}, we consider the local sensitivity of the unnormalized variance estimate, denoting by $g$, i.e., $g(D) = 2N^2 \cdot \hat{\mathbb{V}}[\hat{\tau}]$. Thus, we have $\mathrm{LS}_{\hat{\mathbb{V}}[\hat{\tau}]}(D) = \frac{1}{2N^2} \mathrm{LS}_g(D)$.
We also upper bound $\mathrm{LS}_g(D) \leq \mathrm{LS}^{+}_g(D) + \max_{D^{\prime\prime}: d_{ar}^+(D,D^{\prime\prime})\leq 1} \mathrm{LS}^{-}_g(D^{\prime\prime})$.

We first consider $\mathrm{LS}^{+}_g(D)$.
We write the neighboring dataset $D^\prime = \{(W_i^\prime, Y_i^{\mathrm{obs}\prime}, X_i^\prime)\}_{i=1}^{N+1}$ and also define $T^\prime_x$, $C^\prime_x$, and $L^\prime_i$ accordingly.
W.l.o.g., $D_{N+1}\in D^\prime$ is the added individual data.
Let $x$ be $x=X^\prime_{N+1}$. 

Here we assume w.l.o.g. $W_{N+1} = 1$. When $C_x,T_x$ are non-empty, it holds that
\begin{align*}
    |g(D)-g(D^\prime)| &= |\sum_{i\in T_x \cup C_x} (1+L_i)^2 (\hat{Y}_i(1)-\hat{Y}_i(0))^2 - \sum_{i\in T_x^\prime \cup C_x^\prime} (1+L_i^\prime)^2 (\hat{Y}_i^\prime(1)-\hat{Y}_i^\prime(0))^2| \\
    &= |\sum_{i\in T_x \cup C_x} \left\{(1+L_i)^2 (\hat{Y}_i(1)-\hat{Y}_i(0))^2 - (1+L_i^\prime)^2 (\hat{Y}_i^\prime(1)-\hat{Y}_i^\prime(0))^2 \right\}\\
    &\quad- (1+L_{N+1}^\prime)^2 (\hat{Y}_{N+1}^\prime(1)-\hat{Y}_{N+1}^\prime(0))^2| \\
    &\leq B^2 (\sum_{i\in T_x \cup C_x}\{(1+L_i)^2+(1+L_i^\prime)^2\}+(1+L_{N+1}^\prime)^2)\\
    &\leq B^2 (|T_x|((1+\lceil\frac{|C_x|}{|T_x|}\rceil)^2+(1+\lceil\frac{|C_x^\prime|}{|T_x^\prime|}\rceil)^2)
    +|C_x|((1+\lceil\frac{|T_x|}{|C_x|}\rceil)^2+(1+\lceil\frac{|T_x^\prime|}{|C_x^\prime|}\rceil)^2)\\
    &\quad +(1+\lceil\frac{|C_x^\prime|}{|T_x^\prime|}\rceil)^2)\\
    &\leq B^2 (2|T_x|((1+\lceil\frac{|C_x|}{|T_x|}\rceil)^2 
    + 2|C_x|((1+\lceil\frac{|T_x|+1}{|C_x|}\rceil)^2
    +(1+\lceil\frac{|C_x|}{|T_x|+1}\rceil)^2)
\end{align*}

When $C_x$ is empty, $|g(D)-g(D^\prime)| = 0$.
When $T_x$ is empty and $C_x$ is not empty, $|g(D)-g(D^\prime)| \leq B^2((1+|C_x|)^2+|C_x|(1+1)^2) = B^2((1+|C_x|)^2+4|C_x|)$.

Thus, by symmetry, we have the following upper bound on $\mathrm{LS}^{+}_g(D)$.
\begin{align*}
    \mathrm{LS}^{+}_g(D) \leq \max_x \begin{cases}
        0 & |T_x| = |C_x| = 0\\
        B^2((1+|C_x|)^2+4|C_x|) & |T_x| =  0 \land |C_x| > 0 \\
        B^2((1+|T_x|)^2+4|T_x|)  & |T_x| >  0 \land |C_x| = 0 \\
        B^2\max(2|T_x|((1+\lceil\frac{|C_x|}{|T_x|}\rceil)^2 
    + 2|C_x|((1+\lceil\frac{|T_x|+1}{|C_x|}\rceil)^2
    +(1+\lceil\frac{|C_x|}{|T_x|+1}\rceil)^2,\\
        2|C_x|((1+\lceil\frac{|T_x|}{|C_x|}\rceil)^2 
    + 2|T_x|((1+\lceil\frac{|C_x|+1}{|T_x|}\rceil)^2
    +(1+\lceil\frac{|T_x|}{|C_x|+1}\rceil)^2) & o.w.
    \end{cases}
\end{align*}

Similarly, we have the following.
\begin{align*}
    \mathrm{LS}^{-}_g(D) \leq \max_x \begin{cases}
        0 & |T_x| = 0 \lor |C_x| = 0\\
        B^2\max(2(1+|T_x|)((1+\lceil\frac{|C_x|}{|T_x|}\rceil)^2 
    + 2|C_x|((1+\lceil\frac{|T_x|+1}{|C_x|}\rceil)^2
    +(1+\lceil\frac{|C_x|}{|T_x|}\rceil)^2,\\
        2(1+|C_x|)((1+\lceil\frac{|T_x|}{|C_x|}\rceil)^2 
    + 2|T_x|((1+\lceil\frac{|C_x|+1}{|T_x|}\rceil)^2
    +(1+\lceil\frac{|T_x|}{|C_x|}\rceil)^2) & o.w.
    \end{cases}
\end{align*}

Therefore, 
\begin{align*}
    &\max_{D^{\prime\prime}: d_{ar}^+(D,D^{\prime\prime})\leq 1} \mathrm{LS}^{-}_g(D^{\prime\prime}) \\
    &= \max_x \begin{cases} 0 & |T_x| = |C_x| = 0\\
    B^2\max(8(1+|C_x|) + 2(2+|C_x|)^2 + 4,\\
    4(1+|C_x|)^2 + 2|C_x|((1+\lceil\frac{2}{|C_x|}\rceil)^2 +(1+|C_x|)^2) 
    & |T_x| = 0 \land |C_x| > 0 \\
    B^2\max(8(1+|T_x|) + 2(2+|T_x|)^2 + 4,\\
    4(1+|T_x|)^2 + 2|T_x|((1+\lceil\frac{2}{|T_x|}\rceil)^2 +(1+|T_x|)^2) 
    & |T_x| > 0 \land |C_x| = 0 \\
    B^2\max(
    2(1+|C_x|)(1+\lceil\frac{1+|T_x|}{|C_x|}\rceil)^2+2(1+|T_x|)(1+\lceil\frac{1+|C_x|}{1+|T_x|}\rceil)^2+(1+\lceil\frac{1+|T_x|}{|C_x|}\rceil)^2,\\
    2(2+|T_x|)(1+\lceil\frac{|C_x|}{1+|T_x|}\rceil)^2+2|C_x|(1+\lceil\frac{2+|T_x|}{|C_x|}\rceil)^2+(1+\lceil\frac{|C_x|}{1+|T_x|}\rceil)^2,\\
    2(1+|T_x|)(1+\lceil\frac{1+|C_x|}{|T_x|}\rceil)^2+2(1+|C_x|)(1+\lceil\frac{1+|T_x|}{1+|C_x|}\rceil)^2+(1+\lceil\frac{1+|C_x|}{|T_x|}\rceil)^2,\\
    2(2+|C_x|)(1+\lceil\frac{|T_x|}{1+|C_x|}\rceil)^2+2|T_x|(1+\lceil\frac{2+|C_x|}{|T_x|}\rceil)^2+(1+\lceil\frac{|T_x|}{1+|C_x|}\rceil)^2)
    & o.w.
    \end{cases}
\end{align*}

Combining above, we have the upper bound on $\mathrm{LS}_{\hat{\mathbb{V}}[\hat{\tau}]}(D)$ as below:
\begin{align*}
    \mathrm{LS}_{\hat{\mathbb{V}}[\hat{\tau}]}(D)\leq \frac{1}{2N^2} \left(\mathrm{LS}^{+}_g(D) + \max_{D^{\prime\prime}: d_{ar}^+(D,D^{\prime\prime})\leq 1} \mathrm{LS}^{-}_g(D^{\prime\prime})\right).
\end{align*}

The $\beta$-smooth sensitivity is by definition:
\begin{align*}
    S^*_{\hat{\mathbb{V}}[\hat{\tau}],\beta}(D) = \max_{k=0,\ldots, N} e^{-k\beta} \max_{D^\prime:d(D,D^\prime)\leq k}\mathrm{LS}_{\hat{\mathbb{V}}[\hat{\tau}]}(D^\prime).
\end{align*}
Since the local sensitivity depends only on $|T_x|$ and $|C_x|$, it remains to consider all possible $|T_x^\prime|$'s and $|C_x^\prime|$'s such that $d(D,D^\prime)\leq k$ and compute the local sensitivity for all $k$ and take the maximum.

\bibliographystyle{abbrv}
\bibliography{supplement}